\documentclass[10pt,twocolumn,letterpaper]{article}
\usepackage{cvpr}
\usepackage{times}
\usepackage{epsfig}
\usepackage{graphics}
\usepackage{booktabs}
\usepackage{amsmath}
\usepackage{amssymb}
\usepackage{algorithm}
\usepackage{algorithmic}
\usepackage{bm}
\usepackage{subfigure}
\usepackage{wrapfig}
 \usepackage{url}
\newtheorem{theorem1}{Theorem}
\newcommand{\hdistance}{\mathcal{H}\Delta\mathcal{H}}

\newcommand{\tblcaption}[1]{\def\@captype{table}\caption{#1}}
\newcommand\subcaption[1]{\begin{center}#1\end{center}}

\usepackage{authblk}
\newcommand*{\affaddr}[1]{#1} 
\newcommand*{\affmark}[1][*]{\textsuperscript{#1}}
\newcommand*{\email}[1]{\texttt{#1}}

\usepackage[pagebackref=true,breaklinks=true,letterpaper=true,colorlinks,bookmarks=false]{hyperref}

 \cvprfinalcopy 

\def\cvprPaperID{1470} 

\ifcvprfinal\fi
\begin{document}

\title{Maximum Classifier Discrepancy for Unsupervised Domain Adaptation}
\author{%
  Kuniaki Saito\affmark[1], Kohei Watanabe\affmark[1], Yoshitaka Ushiku\affmark[1], and Tatsuya Harada\affmark[1,2]\\
  \affaddr{\affmark[1]The University of Tokyo}, \affaddr{\affmark[2]RIKEN}\\
  \email{\{k-saito,watanabe,ushiku,harada\}@mi.t.u-tokyo.ac.jp}\\
  }

\maketitle

\begin{abstract}
In this work, we present a method for unsupervised domain adaptation. 
Many adversarial learning methods train domain classifier networks to distinguish the features as either a source or target and train a feature generator network to mimic the discriminator.
Two problems exist with these methods.
First, the domain classifier only tries to distinguish the features as a source or target and thus does not consider task-specific decision boundaries between classes. Therefore, a trained generator can generate ambiguous features near class boundaries.
Second, these methods aim to completely match the feature distributions between different domains, which is difficult because of each domain's characteristics.

To solve these problems, we introduce a new approach that attempts to align distributions of source and target by utilizing the task-specific decision boundaries. 
We propose to maximize the discrepancy between two classifiers' outputs to detect target samples that are far from the support of the source. A feature generator learns to generate target features near the support to minimize the discrepancy. 
Our method outperforms other methods on several datasets of image classification and semantic segmentation. The codes are available at \url{https://github.com/mil-tokyo/MCD_DA}

\end{abstract}
\vspace{-3mm}
\section{Introduction}
\vspace{-2mm}
  \begin{figure}[t]
    \begin{center}
      \includegraphics[width=\hsize]{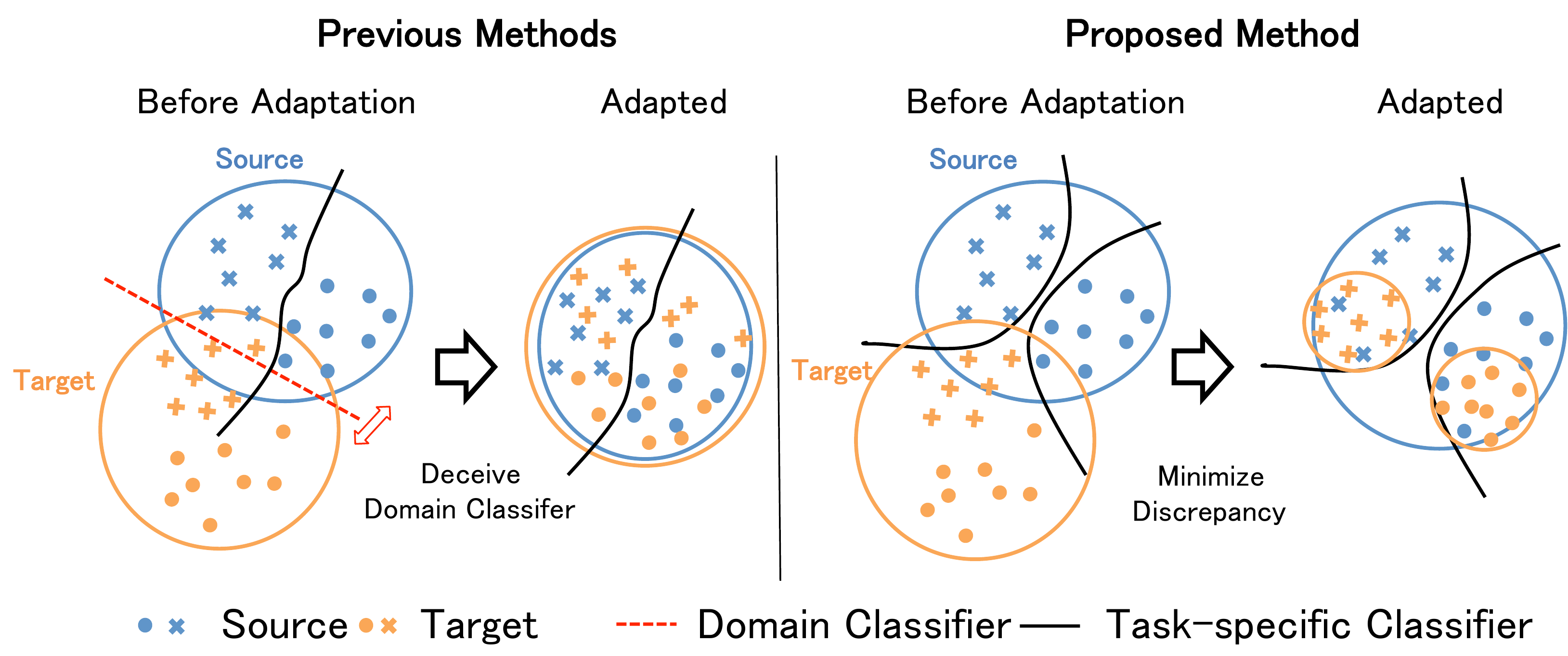}
           \caption{(Best viewed in color.) Comparison of previous and the proposed distribution matching methods.. {\bf Left}: Previous methods try to match different distributions by mimicing the domain classifier. They do not consider the decision boundary. {\bf Right}: Our proposed method attempts to detect target samples outside the support of the source distribution using task-specific classifiers.}
              \label{fig:intro}
    \end{center}
   \end{figure}
The classification accuracy of images has improved substantially with the advent of deep convolutional neural networks (CNN) which utilize numerous labeled samples~\cite{krizhevsky2012imagenet}. However, collecting numerous labeled samples in various domains is expensive and time-consuming.

Domain adaptation (DA) tackles this problem by transferring knowledge from a label-rich domain (i.e., source domain) to a label-scarce domain (i.e., target domain). DA aims to train a classifier using source samples that generalize well to the target domain. However, each domain's samples have different characteristics, which makes the problem difficult to solve. Consider neural networks trained on labeled source images collected from the Web. Although such neural networks perform well on the source images, correctly recognizing target images collected from a real camera is difficult for them. This is because the target images can have different characteristics from the source images, such as change of light, noise, and angle in which the image is captured.
Furthermore, regarding unsupervised DA (UDA), we have access to labeled source samples and only unlabeled target samples. We must construct a model that works well on target samples despite the absence of their labels during training. UDA is the most challenging situation, and we propose a method for UDA in this study.

Many UDA algorithms, particularly those for training neural networks, attempt to match the distribution of the source features with that of the target without considering the category of the samples~\cite{ganin2016domain,sun2016deep,bousmalis2016domain,tzeng2014deep}.
In particular, domain classifier-based adaptation algorithms have been applied to many tasks~\cite{ganin2016domain,bousmalis2016domain}. The methods utilize two players to align distributions in an adversarial manner: domain classifier (i.e., a discriminator) and feature generator. Source and target samples are input to the same feature generator. Features from the feature generator are shared by the discriminator and a task-specific classifier.
The discriminator is trained to discriminate the domain labels of the features generated by the generator whereas the generator is trained to fool it. The generator aims to match distributions between the source and target because such distributions will mimic the discriminator. They assume that such target features are classified correctly by the task-specific classifier because they are aligned with the source samples.

However, this method should fail to extract discriminative features because it does not consider the relationship between target samples and the task-specific decision boundary when aligning distributions. As shown in the left side of Fig. \ref{fig:intro}, the generator can generate ambiguous features near the boundary because it simply tries to make the two distributions similar.

To overcome both problems, we propose to align distributions of features from source and target domain by using the classifier's output for the target samples. 

We introduce a new adversarial learning method that utilizes two types of players: task-specific classifiers and a feature generator. {\it task-specific classifiers} denotes the classifiers trained for each task such as object classification or semantic segmentation. 
Two classifiers take features from the generator.
Two classifiers try to classify source samples correctly and, simultaneously, are trained to detect the target samples that are far from the support of the source. The samples existing far from the support do not have discriminative features because they are not clearly categorized into some classes. Thus, our method utilizes the task-specific classifiers as a discriminator. 
Generator tries to fool the classifiers. In other words, it is trained to generate target features near the support while considering classifiers' output for target samples.
Thus, our method allows the generator to generate discriminative features for target samples because it considers the relationship between the decision boundary and target samples.
This training is achieved in an adversarial manner. In addition, please note that we do not use domain labels in our method.

We evaluate our method on image recognition and semantic segmentation. In many settings, our method outperforms other methods by a large margin.
The contributions of our paper are summarized as follows:

\begin{itemize}
\item We propose a novel adversarial training method for domain adaptation that tries to align the distribution of a target domain by considering task-specific decision boundaries. 
\item We confirm the behavior of our method through a toy problem.
\item We extensively evaluate our method on various tasks: digit classification, object classification, and semantic segmentation.
\end{itemize}
\vspace{-4mm}
\section{Related Work}
\vspace{-1mm}
Training CNN for DA can be realized through various strategies. Ghifary \etal proposed using an autoencoder for the target domain to obtain domain-invariant features~\cite{ghifary2016deep}. Sener \etal proposed using clustering techniques and pseudo-labels to obtain discriminative features~\cite{sener2016learning}. Taigman \etal proposed cross-domain image translation methods~\cite{taigman2016unsupervised}. Matching distributions of the middle features in CNN is considered to be effective in realizing an accurate adaptation. To this end, numerous methods have been proposed~\cite{ganin2016domain,sun2016deep,bousmalis2016domain,purushotham2017variational,tzeng2014deep,sun2015return}.

\begin{figure*}[t]
    \begin{center}
      \includegraphics[width=0.85\hsize]{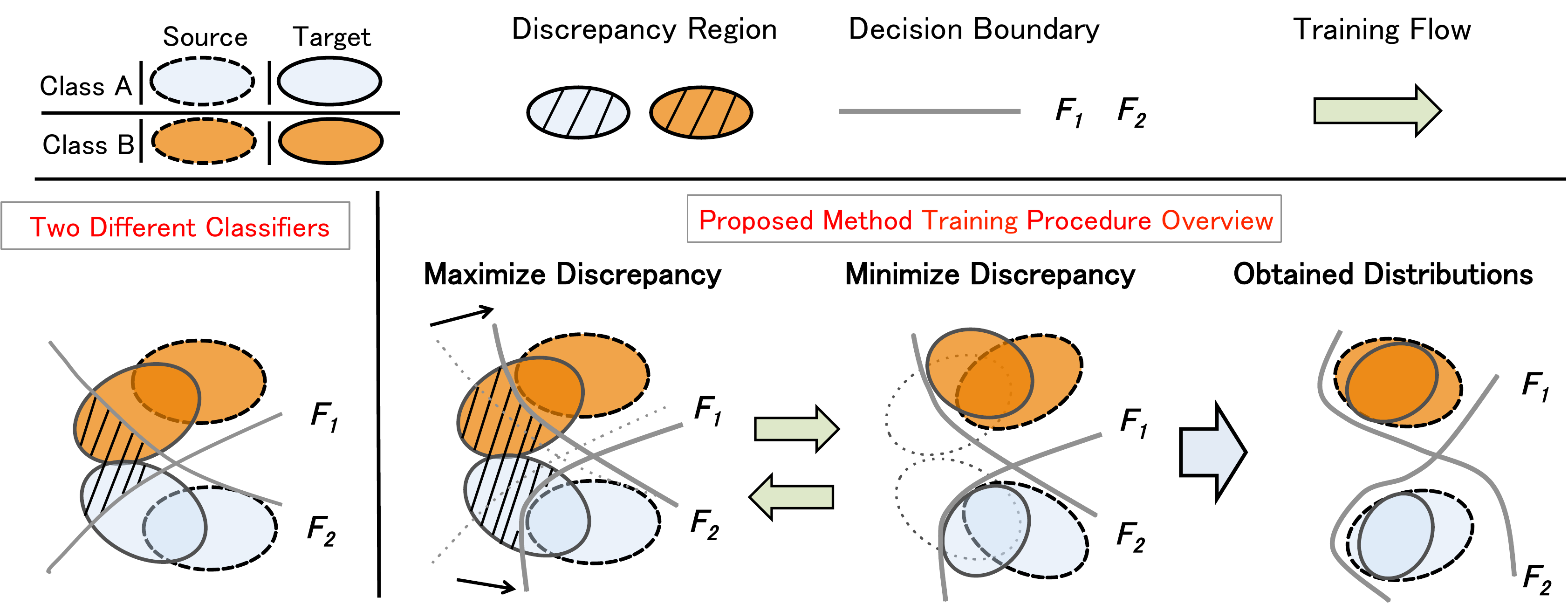}
           \caption{(Best viewed in color.) Example of two classifiers with an overview of the proposed method. Discrepancy refers to the disagreement between the predictions of two classifiers. First, we can see that the target samples outside the support of the source can be measured by two different classifiers (Leftmost, {\it Two different classifiers}).
 Second, regarding the training procedure, we solve a minimax problem in which we find two classifiers that {\it maximize} the discrepancy on the target sample, and then generate features that {\it minimize} this discrepancy.}
              \label{fig:propose}
    \end{center}
 \end{figure*}
The representative method of distribution matching involves training a domain classifier using the middle features and generating the features that deceive the domain classifier~\cite{ganin2016domain}. This method utilizes the techniques used in generative adversarial networks~\cite{GAN}. The domain classifier is trained to predict the domain of each input, and the category classifier is trained to predict the task-specific category labels. Feature extraction layers are shared by the two classifiers. The layers are trained to correctly predict the label of source samples as well as to deceive the domain classifier.
Thus, the distributions of the middle features of the target and source samples are made similar.
Some methods utilize maximum mean discrepancy (MMD)~\cite{long2016unsupervised,long2015learning}, which can be applied to measure the divergence in high-dimensional space between different domains. This approach can train the CNN to simultaneously minimize both the divergence and category loss for the source domain. These methods are based on the theory proposed by~\cite{ben2007analysis}, which states that the error on the target domain is bounded by the divergence of the distributions.
To our understanding, these distribution aligning methods using GAN or MMD do not consider the relationship between target samples and decision boundaries. To tackle these problems, we propose a novel approach using task-specific classifiers as a discriminator.

Consensus regularization is a technique used in multi-source domain adaptation and multi-view learning, in which multiple classifiers are trained to maximize the consensus of their outputs~\cite{luo2008transfer}. 
In our method, we address a training step that minimizes the consensus of two classifiers, which is totally different from consensus regularization. Consensus regularization utilizes samples of multi-source domains to construct different classifiers as in~\cite{luo2008transfer}. In order to construct different classifiers, it relies on the different characteristics of samples in different source domains. By contrast, our method can construct different classifiers from only one source domain.
\section{Method}
In this section, we present the detail of our proposed method. First, we give the overall idea of our method in Section \ref{mtd:overall}. Second, we explain about the loss function we used in experiments in Section \ref{mtd:loss}. Finally, we explain the entire training procedure of our method in Section \ref{mtd:steps}. 
\subsection{Overall Idea}\label{mtd:overall}
We have access to a labeled source image $\mathbf{x_{s}}$ and a corresponding label $y_{s}$ drawn from a set of labeled source images \{$X_{s},Y_{s}$\},
as well as an unlabeled target image $\mathbf{x_{t}}$ drawn from unlabeled target images $X_{t}$.
We train a feature generator network $G$, which takes inputs $\mathbf{x_{s}}$ or $\mathbf{x_{t}}$, and classifier networks $F_1$ and $F_2$, which take features from $G$. 
$F_1$ and $F_2$ classify them into $K$ classes, that is, they output a $K$-dimensional vector of logits. We obtain class probabilities by applying the softmax function for the vector. We use the notation $p_1(\mathbf{y}|\mathbf{x})$, $p_2(\mathbf{y}|\mathbf{x})$ to denote the $K$-dimensional probabilistic outputs for input $\mathbf{x}$ obtained by $F_1$ and $F_2$ respectively.

The goal of our method is to align source and target features by utilizing the task-specific classifiers as a discriminator in order to consider the relationship between class boundaries and target samples. For this objective, we have to detect target samples far from the support of the source.
The question is how to detect target samples far from the support. These target samples are likely to be misclassified by the classifier learned from source samples because they are near the class boundaries.
Then, in order to detect these target samples, we propose to utilize the disagreement of the two classifiers on the prediction for target samples. Consider two classifiers ($F_1$ and $F_2$) that have different characteristics in the leftmost side of Fig. \ref{fig:propose}. We assume that the two classifiers can classify source samples correctly. This assumption is realistic because we have access to labeled source samples in the setting of UDA. In addition, please note that $F_1$ and $F_2$ are initialized differently to obtain different classifiers from the beginning of training.
Here, we have the key intuition that target samples outside the support of the source are likely to be classified differently by the two distinct classifiers. This region is denoted by black lines in the leftmost side of Fig. \ref{fig:propose} ({\it Discrepancy Region}). Conversely, if we can measure the disagreement between the two classifiers and train the generator to minimize the disagreement, the generator will avoid generating target features outside the support of the source.
Here, we consider measuring the difference for a target sample using the following equation, $d(p_1(\mathbf{y}|\mathbf{x_t}),p_2(\mathbf{y}|\mathbf{x_t}))$ where $d$ denotes the function measuring divergence between two probabilistic outputs. This term indicates how the two classifiers disagree on their predictions and, hereafter, we call the term as {\it discrepancy}. Our goal is to obtain a feature generator that can minimize the discrepancy on target samples.

In order to effectively detect target samples outside the support of the source, we propose to train discriminators ($F_1$ and $F_2$) to maximize the discrepancy given target features ({\it Maximize Discrepancy} in Fig. \ref{fig:propose}). Without this operation, the two classifiers can be very similar ones and cannot detect target samples outside the support of the source.
We then train the generator to fool the discriminator, that is, by minimizing the discrepancy ({\it Minimize Discrepancy} in Fig. \ref{fig:propose}).
This operation encourages the target samples to be generated inside the support of the source. This adversarial learning steps are repeated in our method. Our goal is to obtain the features, in which the support of the target is included by that of the source ({\it Obtained Distributions} in Fig. \ref{fig:propose}). We show the loss function used for discrepancy loss in the next section. Then, we detail the training procedure.

\subsection{Discrepancy Loss}\label{mtd:loss}
In this study, we utilize the absolute values of the difference between the two classifiers' probabilistic outputs as discrepancy loss:
\begin{equation}
 d(p_1,p_2) = \frac{1}{K}\sum_{k=1}^{K}|{p_1}_{k}-{p_2}_{k}|,
  \end{equation}
  where the ${p_1}_{k}$ and ${p_2}_{k}$ denote probability output of ${p_1}$ and ${p_2}$ for class $k$ respectively. The choice for L1-distance is based on the Theorem \label{th:th_1}. Additionally, we experimentally found that L2-distance does not work well. 

\subsection{Training Steps}\label{mtd:steps}
To sum up the previous discussion in Section \ref{mtd:overall}, we need to train two classifiers, which take inputs from the generator and maximize $d(p_1(\mathbf{y}|\mathbf{x_t}),p_2(\mathbf{y}|\mathbf{x_t}))$, and the generator which tries to mimic the classifiers. Both the classifiers and generator must classify source samples correctly. We will show the manner in which to achieve this.
We solve this problem in three steps.

\newcommand{\mymin}{\mathop{\rm min}\limits}
\newcommand{\mymax}{\mathop{\rm max}\limits}
\newcommand{\1}{\mbox{1}\hspace{-0.25em}\mbox{l}}
\begin{figure}[t]
  \begin{center}
   \includegraphics[width=0.9\hsize]{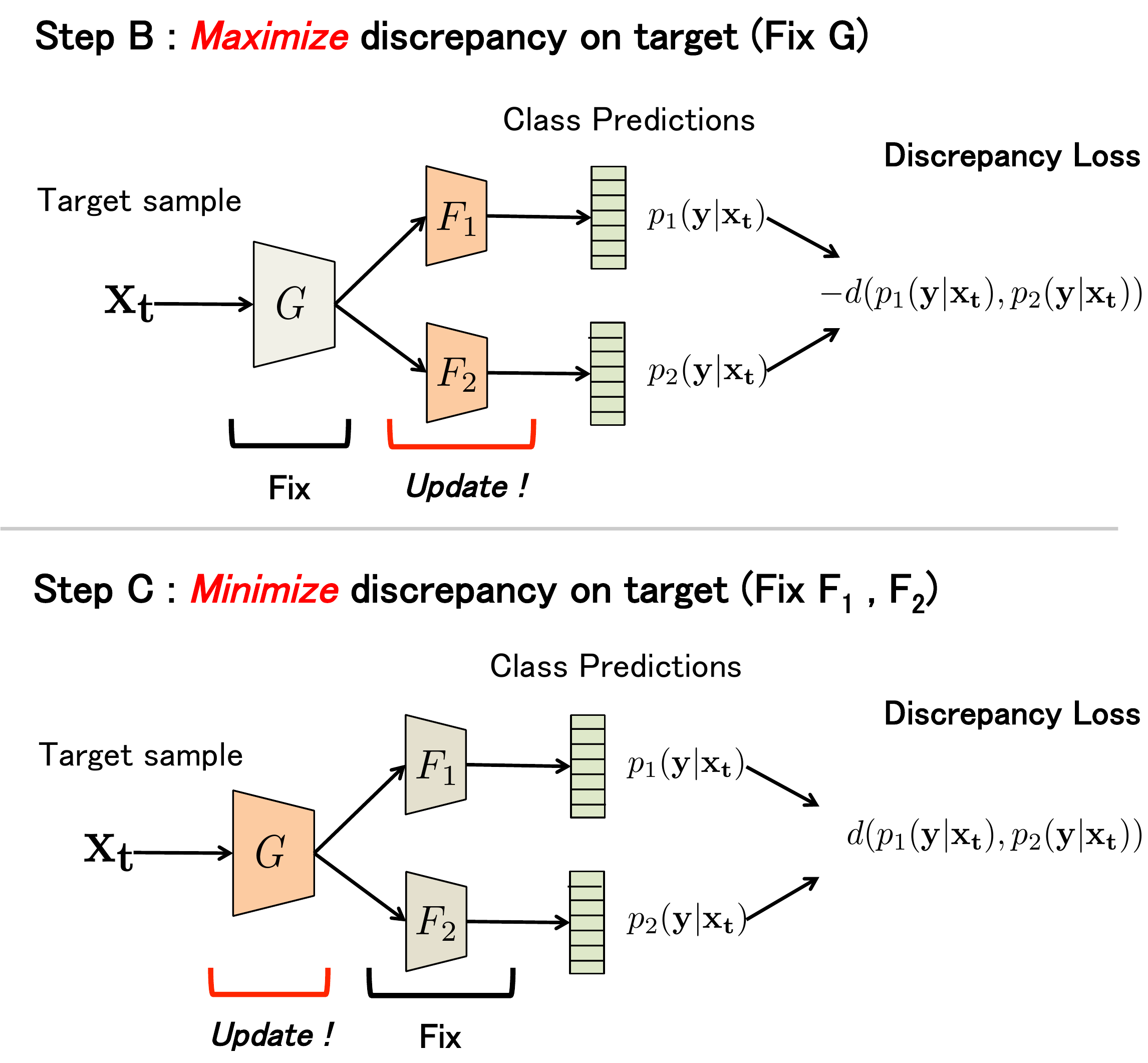}
  \end{center}
 
  \caption{Adversarial training steps of our method. We separate the network into two modules: generator (${\it G}$) and classifiers (${\it F_{1}, F_{2}}$). The classifiers learn to maximize the discrepancy {\bf Step B} on the target samples, and the generator learns to minimize the discrepancy {\bf Step C}. Please note that we employ a training {\bf Step A} to ensure the discriminative features for source samples.}
  \label{fig:model}
\end{figure}
 
\textbf{Step A}
First, we train both classifiers and generator to classify the source samples correctly. In order to make classifiers and generator obtain task-specific discriminative features, this step is crucial. We train the networks to minimize softmax cross entropy. The objective is as follows:
 \begin{equation}
   \mymin_{G,F_1,F_2} \mathcal{L}(X_{s},Y_{s}).
  \end{equation}
  \begin{equation}
   \mathcal{L}(X_{s},Y_{s}) = -{\mathbb{E}_{(\mathbf{x_{s}},y_{s})\sim(X_{s},Y_{s})}}\sum_{k=1}^{K}{\1_{[k=y_{s}]}}\log p({\mathbf y}|{\mathbf x_{s}})
   \label{eq:crossentropy}
  \end{equation}
\textbf{Step B}
In this step, we train the classifiers ($F_{1}, F_{2}$) as a discriminator for a fixed generator ($G$). By training the classifiers to increase the discrepancy, they can detect the target samples excluded by the support of the source. This step corresponds to {\bf Step B} in Fig. \ref{fig:model}. We add a classification loss on the source samples. Without this loss, we experimentally found that our algorithm's performance drops significantly. We use the same number of source and target samples to update the model. The objective is as follows:
\begin{equation}
  \mymin_{F_1,F_2} \mathcal{L}(X_{s},Y_{s}) - \mathcal{L}_{\rm adv}(X_{t}). \\
\end{equation}
\begin{equation}
  \mathcal{L}_{\rm adv}(X_{t}) = {\mathbb{E}_{\mathbf{x_{t}}\sim X_{t}}}[d(p_1(\mathbf{y}|\mathbf{x_t}),p_2(\mathbf{y}|\mathbf{x_t}))]
  \label{eq:sensitivity}
\end{equation}
\textbf{Step C}
We train the generator to minimize the discrepancy for fixed classifiers. This step corresponds to {\bf Step C} in Fig. \ref{fig:model}. The number $n$ indicates the number of times we repeat this for the same mini-batch. This number is a hyper-parameter of our method. This term denotes the trade-off between the generator and the classifiers.
The objective is as follows:
\begin{equation}
 \mymin_{G} \mathcal{L}_{\rm adv}(X_{t}).\\
 \end{equation}
These three steps are repeated in our method. To our understanding, the order of the three steps is not important. Instead, our major concern is to train the classifiers and generator in an adversarial manner under the condition that they can classify source samples correctly.

\subsection{Theoretical Insight}
Since our method is motivated by the theory proposed by Ben-David \etal.~\cite{ben2010theory}, we want to show the relationship between our method and the theory in this section.

Ben-David \etal~\cite{ben2010theory} proposed the theory that bounds the expected error on the target samples, $R_{\mathcal{T}}(h)$, by using three terms: (i) expected error on the source domain, $R_{\mathcal{S}}(h)$; (ii) $\mathcal{H} \Delta \mathcal{H}$-distance ($d_{{\hdistance}}(\mathcal{S},\mathcal{T})$), which is measured as the discrepancy between two classifiers; and (iii) the shared error of the ideal joint hypothesis, $\lambda$. $\mathcal{S}$ and $\mathcal{T}$ denote source and target domain respectively. Another theory~\cite{ben2007analysis} bounds the error on the target domain, which introduced $\mathcal{H}$-distance ($d_{{\mathcal{H}}}(\mathcal{S},\mathcal{T})$) for domain divergence. The two theories and their relationships can be explained as follows.
\begin{theorem1}\label{th:th_1}
  Let $H$ be the hypothesis class. Given two domains $\mathcal{S}$ and $\mathcal{T}$, we have
   \begin{eqnarray}
  \begin{split}
  \forall h  \in H, R_{\mathcal{T}}(h) &\leq R_{\mathcal{S}}(h)  +\frac{1}{2}{d_{\mathcal{H} \Delta \mathcal{H}}(\mathcal{S},\mathcal{T})}+\lambda \\
  &\leq R_{\mathcal{S}}(h)  +\frac{1}{2}{d_{\mathcal{H}}(\mathcal{S},\mathcal{T})}+\lambda \\
  \end{split}
  \label{eq:main}
  \end{eqnarray}
  where
\scriptsize
  \begin{eqnarray*}
    \begin{split}
      d_{{\hdistance}}(\mathcal{S},\mathcal{T}) &= 2\sup_{(h,h{'})\in \mathcal{H}^{2}} \left| \underset{{\bf x}\sim \mathcal{S}}{\mathbf{E}} {\rm I}\bigl[h({\bf x}) \neq h^{'}({\bf x}) \bigr]- \underset{{\bf x}\sim \mathcal{T}}{\mathbf{E}} {\rm I}\bigl[h({\bf x}) \neq h^{'}({\bf x}) \bigr]\right| \\
        d_{{\mathcal{H}}}(\mathcal{S},\mathcal{T}) &= 2\sup_{h\in \mathcal{H}} \left| \underset{{\bf x}\sim \mathcal{S}}{\mathbf{E}} {\rm I} \bigl[h({\bf x}) \neq 1 \bigr] - \underset{{\bf x}\sim \mathcal{T}}{\mathbf{E}} {\rm I} \bigl[h({\bf x}) \neq 1 \bigr]\right|, \\
\lambda&=\min \left[R_{\mathcal{S}}(h)+R_{\mathcal{T}}(h)\right]\\
\end{split}
  \end{eqnarray*}
  \normalsize
  Here, $R_{\mathcal{T}}(h)$ is the error of hypothesis $h$ on the target domain, and $R_{\mathcal{S}}(h)$ is the corresponding error on the source domain. ${\rm I}[a]$ is the indicator function, which is 1 if predicate a is true and 0 otherwise.
\label{th:thm1}
\end{theorem1}
$\mathcal{H}$-distance is shown to be empirically measured by the error of the domain classifier, which is trained to discriminate the domain of features. $\lambda$ is a constant---the shared error of the ideal joint hypothesis---which is considered sufficiently low to achieve an accurate adaptation.
Earlier studies~\cite{ganin2016domain,sun2016deep,bousmalis2016domain,purushotham2017variational,tzeng2014deep} attempted to measure and minimize $\mathcal{H}$-distance in order to realize the adaptation. As this inequality suggests, $\mathcal{H}$-distance upper-bounds the $\mathcal{H} \Delta \mathcal{H}$-distance.
We will show the relationship between our method and $\mathcal{H} \Delta \mathcal{H}$-distance.

Regarding $d_{{\hdistance}}(\mathcal{S},\mathcal{T})$, if we consider that $h$ and $h{'}$ can classify source samples correctly, the term $\scalebox{0.9}{$\displaystyle \underset{{\bf x}\sim \mathcal{S}}{\mathbf{E}} {\rm I}\bigl[h({\bf x}) \neq h^{'}({\bf x}) \bigr]$}$ is assumed to be very low. $h$ and $h{'}$ should agree on their predictions on source samples. Thus, $d_{{\hdistance}}(\mathcal{S},\mathcal{T})$ is approximately calculated as  $\scalebox{0.9}{$\displaystyle \sup_{(h,h{'})\in \mathcal{H}^{2}}\underset{{\bf x}\sim \mathcal{T}}{\mathbf{E}} {\rm I}\bigl[h({\bf x}) \neq h^{'}({\bf x}) \bigr]$}$, which denotes the supremum of the expected disagreement of two classifiers' predictions on target samples.

We assume that $h$ and $h^{'}$ share the feature extraction part. Then, we decompose the hypothesis $h$ into $G$ and $F_1$, and $h^{'}$ into $G$ and $F_2$. $G$, $F_1$ and $F_2$ correspond to the network in our method.
If we substitute these notations into the $\scalebox{0.9}{$\displaystyle \sup_{(h,h{'})\in \mathcal{H}^{2}}\underset{{\bf x}\sim \mathcal{T}}{\mathbf{E}} {\rm I}\bigl[h({\bf x}) \neq h^{'}({\bf x}) \bigr]$}$ and for fixed $G$, the term will become
\begin{equation}
  \scalebox{0.95}{$\displaystyle \sup_{F_1,F_2}\underset{{\bf x}\sim \mathcal{T}}{\mathbf{E}} {\rm I}\left[F_{1}\circ G({\bf x}) \neq F_{2}\circ G({\bf x}) \right]$}\label{eq:eq1}.
\end{equation}
Furthermore, if we replace $\sup$ with $\max$ and minimize the term with respect to $G$, we obtain
\begin{equation}
  \scalebox{0.95}{$\displaystyle \mymin_{G}\mymax_{F_1,F_2} \underset{{\bf x}\sim \mathcal{T}}{\mathbf{E}} {\rm I}\left[F_{1}\circ G({\bf x}) \neq F_{2}\circ G({\bf x}) \right]$}\label{eq:eq2}.
\end{equation}
This equation is very similar to the mini-max problem we solve in our method, in which classifiers are trained to maximize their discrepancy on target samples and generator tries to minimize it. Although we must train all networks to minimize the classification loss on source samples, we can see the connection to the theory proposed by~\cite{ben2010theory}.
 \vspace{-3mm}
\section{Experiments on Classification}
 \vspace{-3mm}
First, we observed the behavior of our model on toy problem.
Then, we performed an extensive evaluation of the proposed methods on the following datasets: digits, traffic signs, and object classification. 

\begin{figure}[t]
    \begin{minipage}[t]{\hsize}
            \subcaption{Comparison of three decision boundaries}
      \centering
      \subfigure[Source Only]{\includegraphics[width=0.3\hsize]{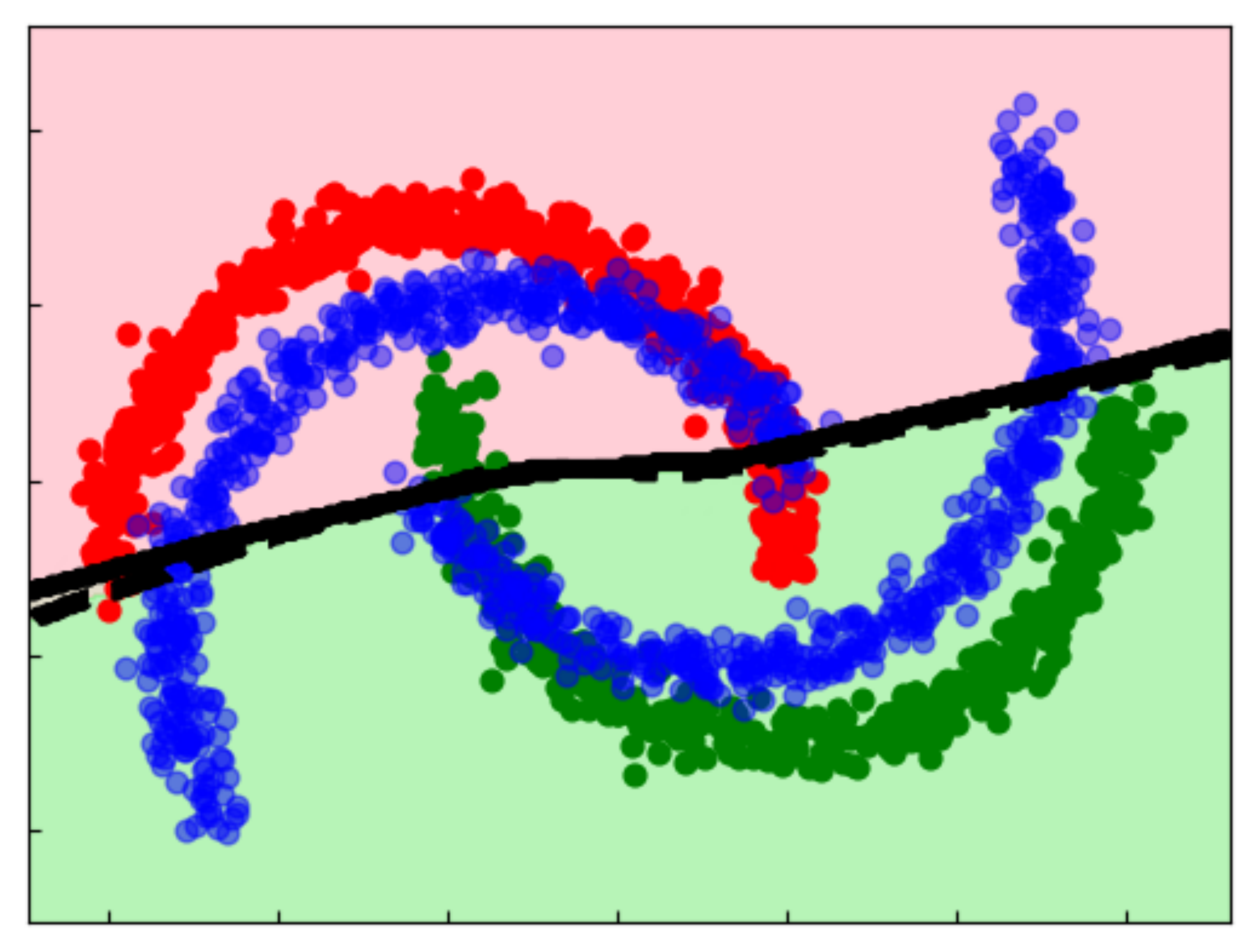}
      \label{fig:rot30_1}}
      \centering
      \subfigure[{\small No Step C}]{\includegraphics[width=0.3\hsize]{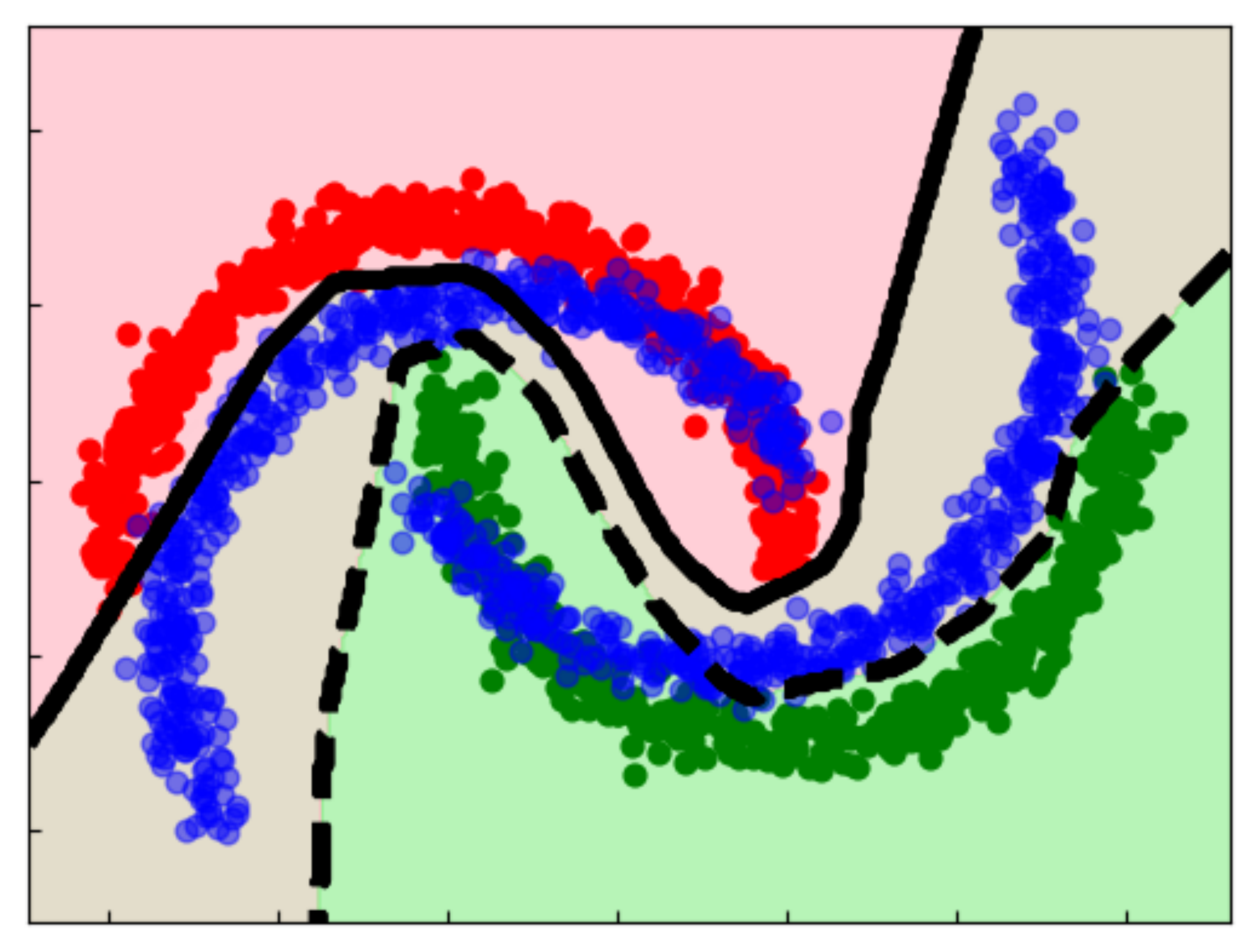}
      \label{fig:rot30_2}}
      \centering
      \subfigure[Proposed]{\includegraphics[width=0.3\hsize]{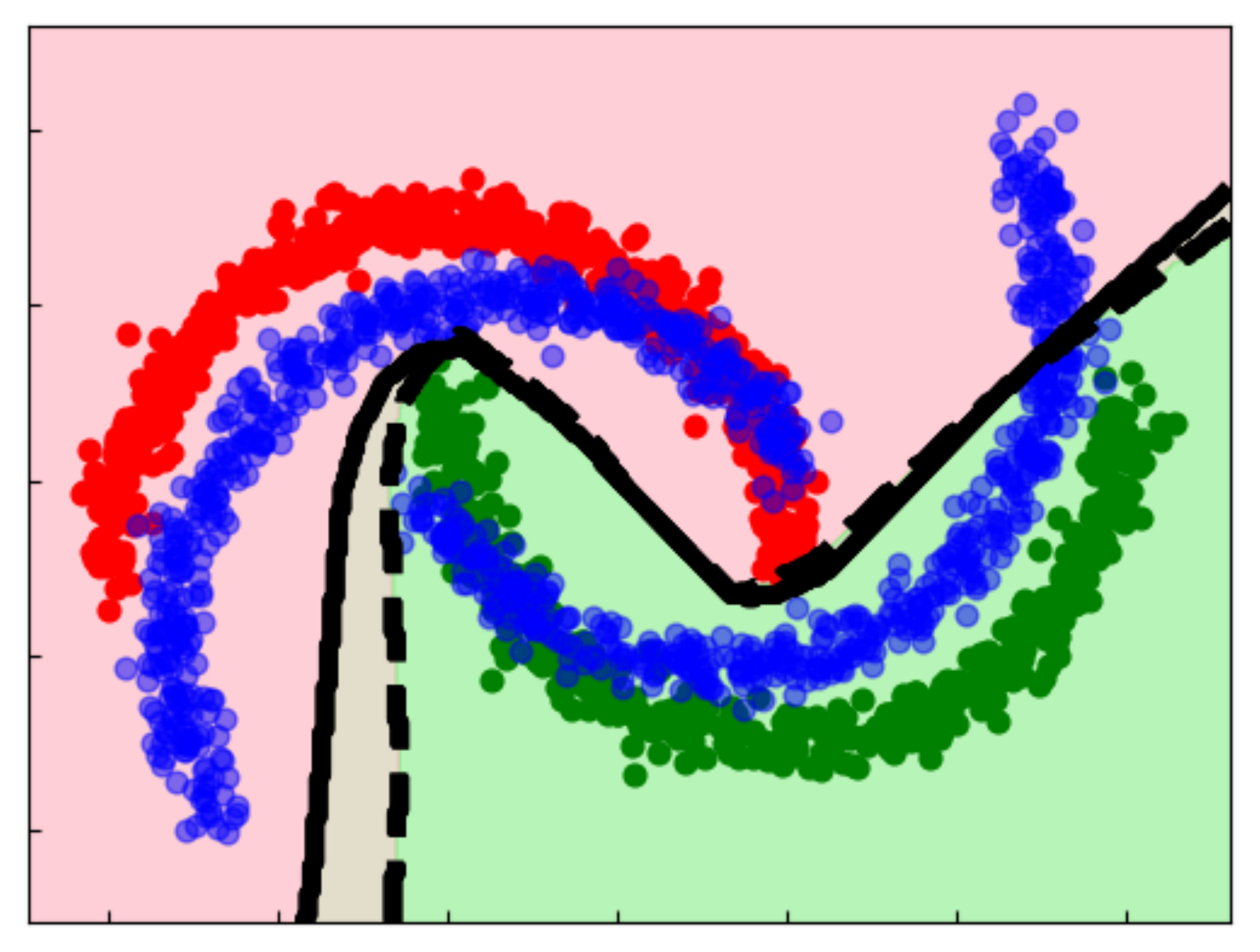}\label{fig:rot30_3}}
    \end{minipage}
 \caption{(Best viewed in color.) Red and green points indicate the source samples of class 0 and 1, respectively. Blue points are target samples generated by rotating source samples. The dashed and normal lines are two decision boundaries in our method. The pink and light green regions are where the results of both classifiers are class 0 and 1, respectively. Fig. \ref{fig:rot30_1} is the model trained only on source samples. Fig. \ref{fig:rot30_2} is the model trained to increase discrepancy of the two classifiers on target samples without using Step C. Fig. \ref{fig:rot30_3} shows our proposed method.}
  \label{fig:rotation}
    \end{figure}

\begin{figure*}[t]
\begin{minipage}{0.3\hsize}
  \centering
  \subfigure[SVHN to MNIST]{\includegraphics[width=0.92\hsize]{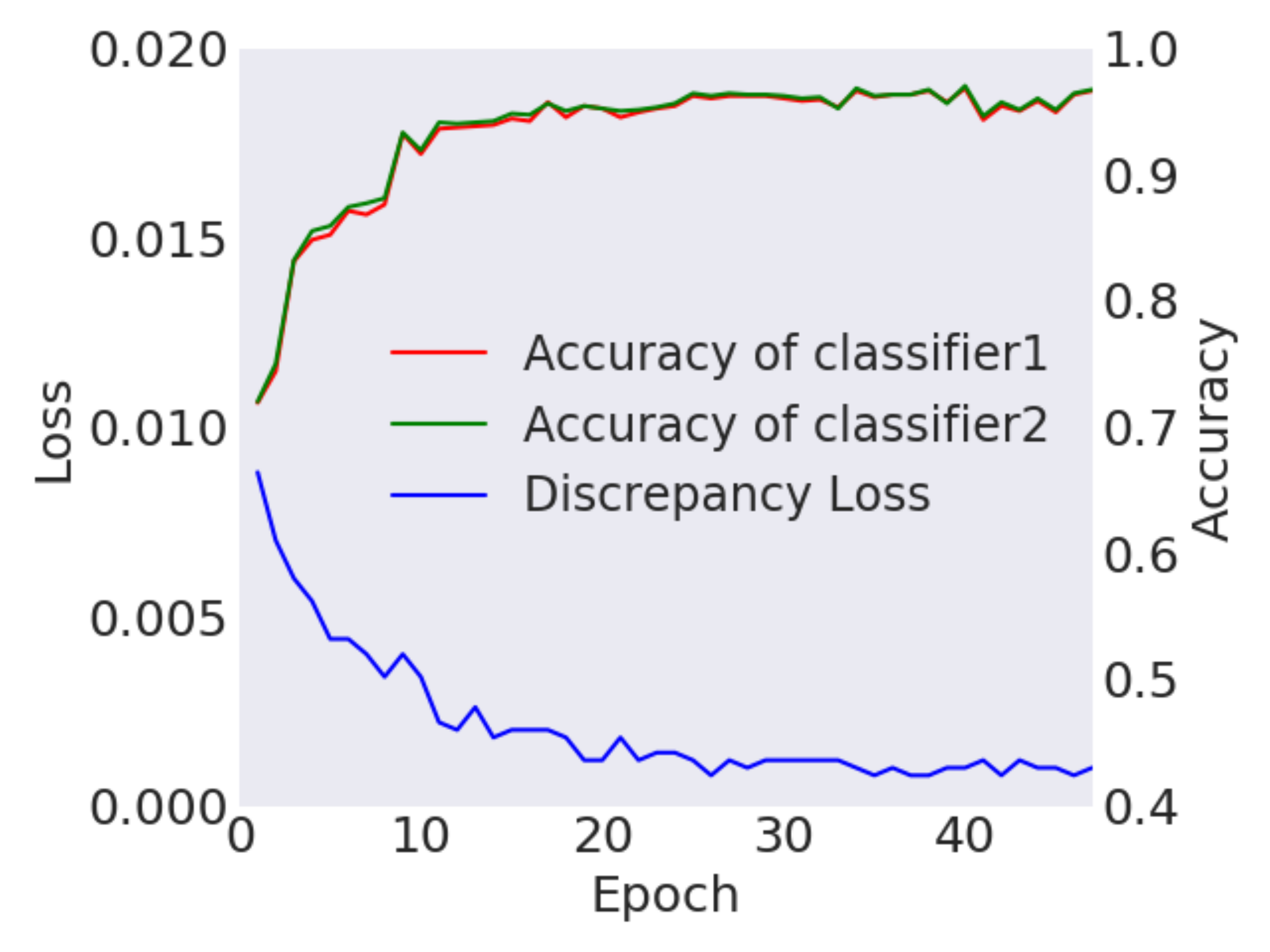}\label{fig:svhn_graph}}
\end{minipage}
\begin{minipage}{0.3\hsize}
  \centering
  \subfigure[SYN SIGN to GTSRB]{\includegraphics[width=0.9\hsize]{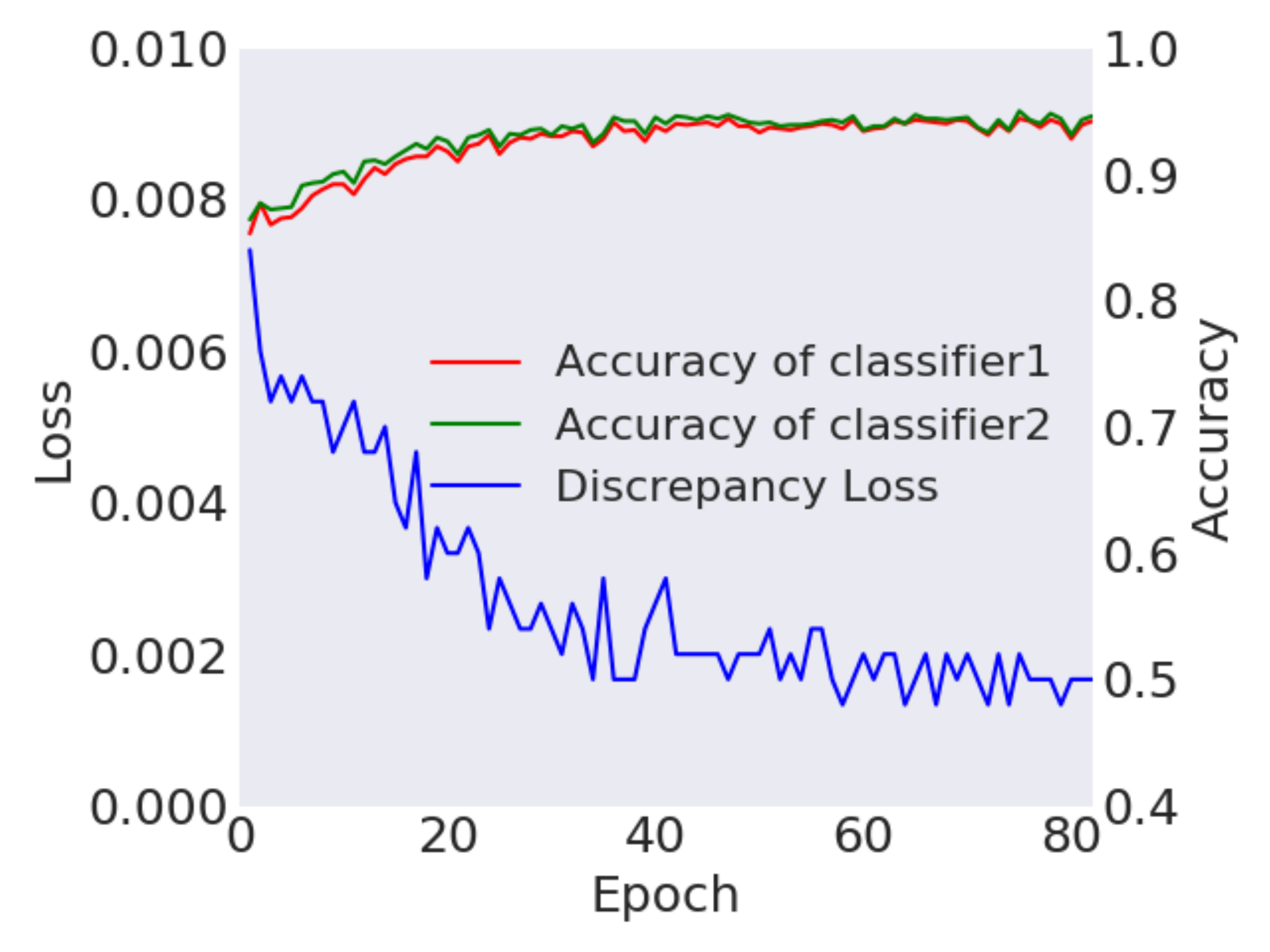}\label{fig:synth_graph}}
\end{minipage}
    \begin{minipage}{0.38\hsize}
  \centering
   \begin{subfigure}[Source Only]{
    \centering
   \includegraphics[width=0.48\hsize,height=0.48\hsize]{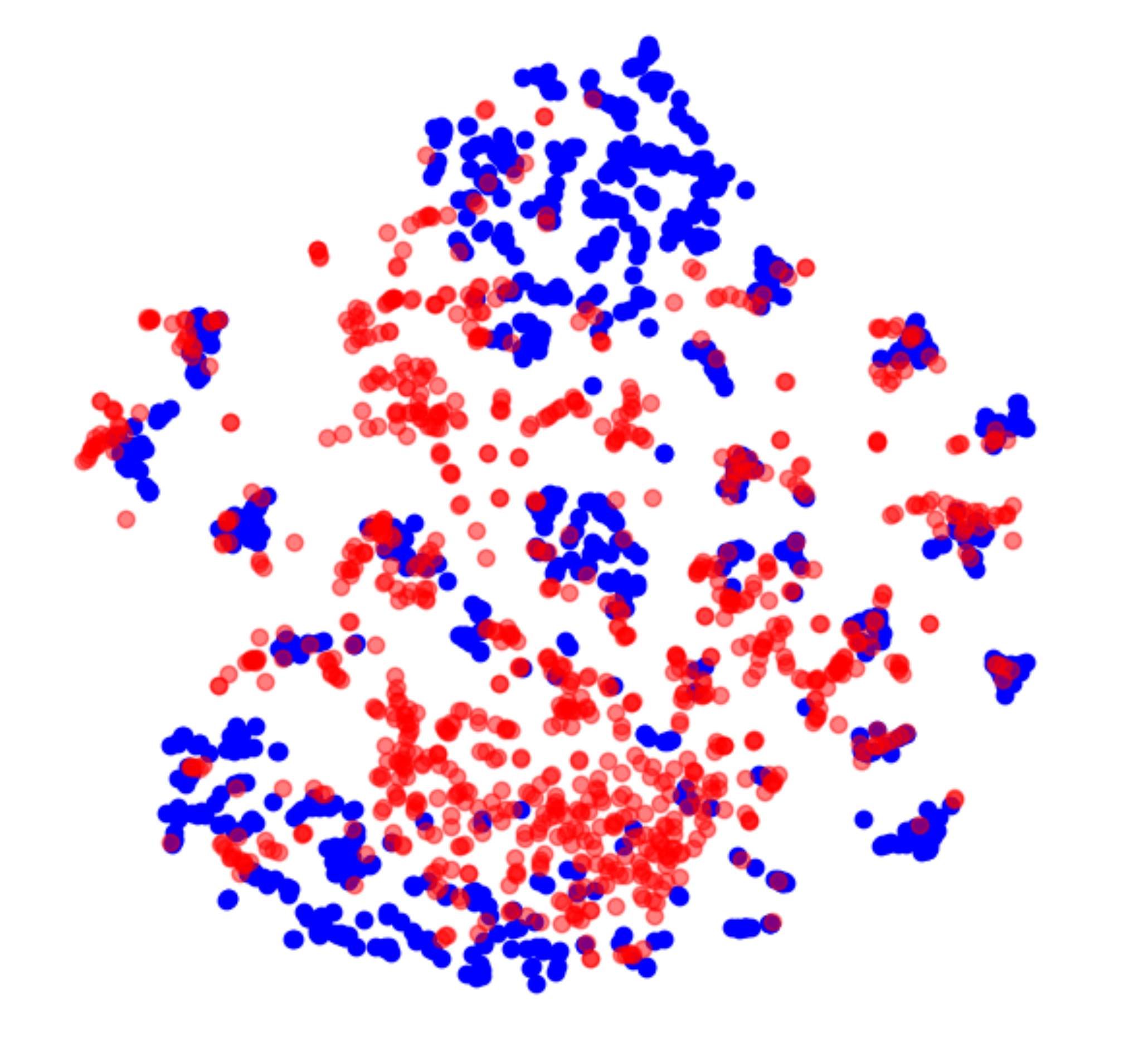}\label{fig:gtsr_noad}}
    \end{subfigure}
      \begin{subfigure}[Adapted (Ours)]{
    \centering
   \includegraphics[width=0.48\hsize,height=0.48\hsize]{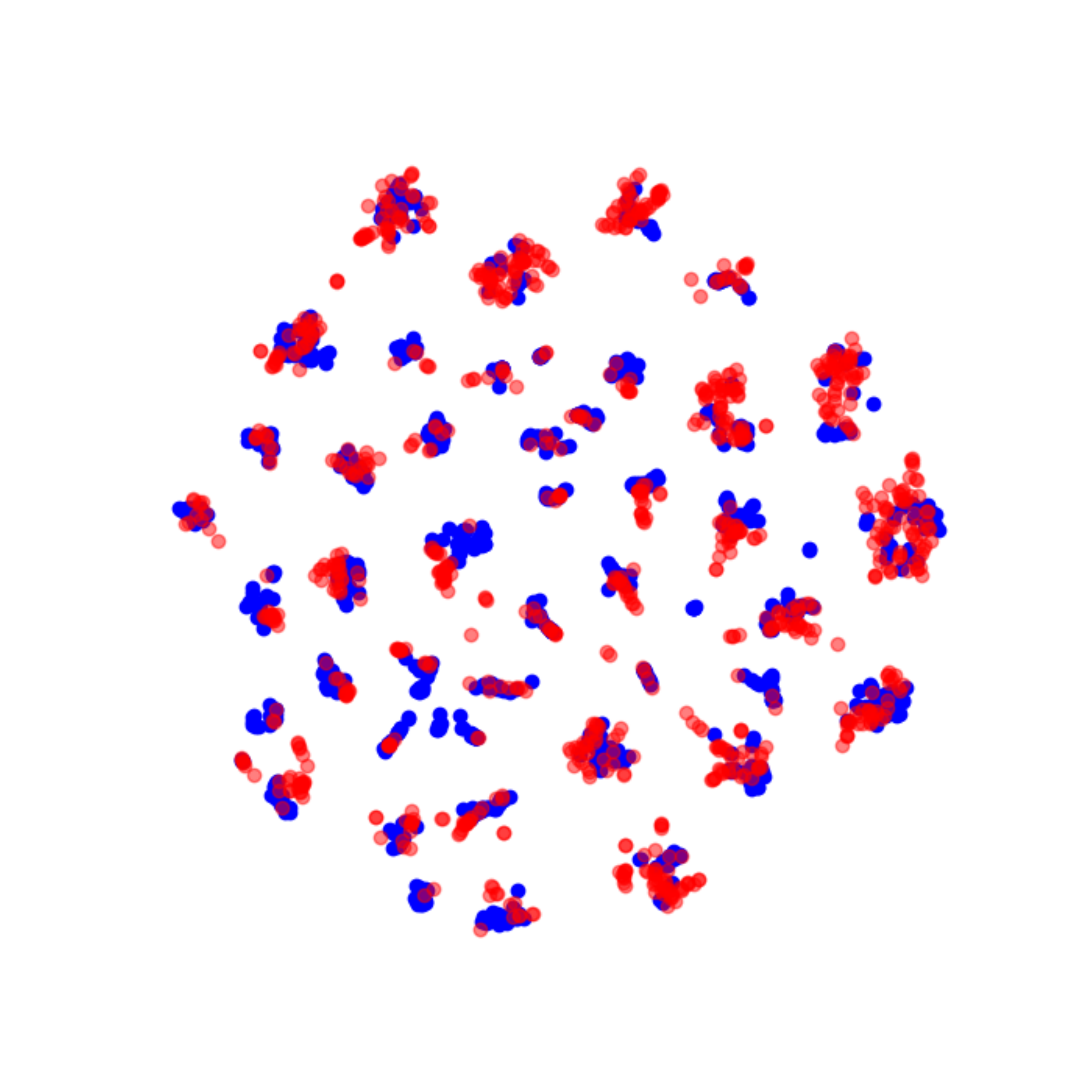}\label{fig:gtsr_ad}}
    \end{subfigure}
    \end{minipage}
  \caption{(Best viewed in color.) {\bf Left}: Relationship between discrepancy loss ({\bf blue} line) and accuracy ({\bf red} and {\bf green} lines) during training. As discrepancy loss decreased, accuracy improved. {\bf Right}: Visualization of features obtained from last pooling layer of the generator in adaptation from SYN SIGNS to GTSRB using t-SNE~\cite{maaten2008visualizing}. {\bf Red} and {\bf blue} points indicate the target and source samples, respectively. All samples are testing samples. We can see that applying our method makes the target samples discriminative.}
  
\end{figure*}

 \subsection{Experiments on Toy Datasets}
  \vspace{-3mm}
In the first experiment, we observed the behavior of the proposed method on {\it inter twinning moons} 2D problems, in which we used {\it scikit-learn}~\cite{pedregosa2011scikit} to generate the target samples by rotating the source samples. The goal of the experiment was to observe the learned classifiers' boundary. For the source samples, we generated a lower moon and an upper moon, labeled 0 and 1, respectively. Target samples were generated by rotating the angle of the distribution of the source samples. We generated 300 source and target samples per class as the training samples. In this experiment, we compared the decision boundary obtained from our method with that obtained from both the model trained only on source samples and from that trained only to increase the discrepancy. In order to train the second comparable model, we simply skipped Step C in Section \ref{mtd:steps} during training. 
We tested the method on 1000 target samples and visualized the learned decision boundary with source and target samples. Other details including the network architecture used in this experiment are provided in our supplementary material.
\begin{table}[t]
\begin{center}
\scalebox{0.7}{
  \begin{tabular}{|l|c|c|c|c|c|c|}
\toprule
& {\bf SVHN}&{\bf SYNSIG}&{\bf MNIST}&{\bf MNIST*}&{\bf USPS}\\
{\bf METHOD}&to&to&to&to&to\\
 & {\bf MNIST} &{\bf GTSRB}&{\bf USPS}&{\bf USPS*}&{\bf MNIST}\\ \hline
Source Only &67.1&85.1&76.7&79.4&63.4\\\hline
\multicolumn{6}{|c|}{{\it Distribution Matching based Methods}}\\\hline
 MMD $\dagger$~\cite{long2015learning}&71.1&91.1&-&81.1&-\\
 DANN $\dagger$~\cite{ganin2014unsupervised}&71.1&88.7&77.1$\pm$1.8&85.1&73.0$\pm$0.2\\
 DSN $\dagger$~\cite{bousmalis2016domain}&82.7&93.1&91.3&-&-\\
  ADDA~\cite{tzeng2017adversarial}&76.0$\pm$1.8&-&89.4$\pm$0.2&-&90.1$\pm$0.8\\
  CoGAN~\cite{liu2016coupled}&-&-&91.2$\pm$0.8&-&89.1$\pm$0.8\\
  PixelDA~\cite{bousmalis2016unsupervised}&-&-&-&95.9&-\\\hline
 Ours ($n=2$)&94.2$\pm$2.6&93.5$\pm$0.4&92.1$\pm$0.8&93.1$\pm$1.9&90.0$\pm$1.4\\
 Ours ($n=3$)& 95.9$\pm$0.5&94.0$\pm$0.4&93.8$\pm$0.8&95.6$\pm$0.9&91.8$\pm$0.9\\
 Ours ($n=4$)&{\bf96.2}$\pm$0.4&{\bf 94.4}$\pm$0.3&{\bf94.2}$\pm$0.7&{\bf96.5}$\pm$0.3&{\bf94.1}$\pm$0.3\\
 \hline
 \multicolumn{6}{|c|}{{\it Other Methods}}\\\hline
 ATDA $\dagger$~\cite{saito2017asymmetric}&86.2&96.2&-&-&-\\
 ASSC~\cite{haeusser2017associative} &95.7$\pm$1.5&82.8$\pm$1.3&-&-&-\\
 DRCN~\cite{ghifary2016deep}&82.0$\pm$0.1&-&91.8$\pm$0.09&-&73.7$\pm$0.04\\
 \bottomrule[1.5pt]
\end{tabular}}
\end{center}
 \caption{Results of the visual DA experiment on the digits and traffic signs datasets. The results are cited from each study. The score of MMD is cited from DSN~\cite{bousmalis2016domain}. Please note that $\dagger$ means that the method used a few labeled target samples as validation, which is different from our setting. We repeated each experiment 5 times and report the average and the standard deviation of the accuracy. The accuracy was obtained from classifier $F_1$. Including the methods that used the labeled target samples for validation, our method achieved good performance. MNIST* and USPS* mean that we used all of the training samples to train the model.}
 \label{tb:exp_digit}
 \end{table}
\begin{table*}[t]
\begin{center}
\label{my-label}
\scalebox{0.85}{
\begin{tabular}{l||cccccccccccc|c}
\toprule[1.5pt]
Method     & plane & bcycl & bus  & car  & horse & knife & mcycl & person & plant & sktbrd & train & truck & mean \\\hline
Source Only & 55.1      & 53.3    & 61.9 & 59.1 & 80.6  & 17.9  & 79.7        & 31.2   & 81.0    & 26.5       & 73.5  & 8.5   & 52.4 \\
MMD~\cite{long2015learning}      & 87.1      & 63.0      & 76.5 & 42.0   & 90.3  & 42.9  & {\bf 85.9}        & 53.1   & 49.7  & 36.3       & {\bf 85.8}  & 20.7  & 61.1 \\
DANN~\cite{ganin2014unsupervised}       & 81.9      & {\bf 77.7}    & 82.8 & 44.3 & 81.2  & 29.5  & 65.1        & 28.6   & 51.9  & {\bf 54.6}       & 82.8  & 7.8   & 57.4 \\\hline
Ours ($n=2$)&81.1&55.3&83.6&{\bf 65.7}&87.6&72.7&83.1&73.9&85.3&47.7&73.2&27.1&69.7\\
Ours ($n=3$)&{\bf 90.3}&49.3&82.1&62.9&{\bf 91.8}&69.4&83.8&72.8&79.8&53.3&81.5&{\bf 29.7}&70.6\\
Ours ($n=4$)& 87.0&60.9&{\bf 83.7}&64.0&88.9&{\bf 79.6}&84.7&{\bf76.9}&{\bf 88.6}&40.3&83.0&25.8&{\bf 71.9}\\
\bottomrule[1.5pt]
\end{tabular}}
\end{center}
  \vspace{-3mm}
\caption{Accuracy of ResNet101 model fine-tuned on the VisDA dataset. The reported accuracy was obtained after 10 epoch updates. }
\label{tb:visda}
\end{table*}
As we expected, when we trained the two classifiers to increase the discrepancy on the target samples, two classifiers largely disagreed on their predictions on target samples (Fig. \ref{fig:rot30_2}). This is clear when compared to the source only model (Fig. \ref{fig:rot30_1}). Two classifiers were trained on the source samples without adaptation, and the boundaries seemed to be nearly the same. Then, our proposed method attempted to generate target samples that reduce the discrepancy. Therefore, we could expect that the two classifiers will be similar. Fig. \ref{fig:rot30_3} demonstrates the assumption. The decision boundaries are drawn considering the target samples. The two classifiers output nearly the same prediction for target samples, and they classified most target samples correctly.
\subsection{Experiments on Digits Datasets}
In this experiment, we evaluate the adaptation of the model on three scenarios. The example datasets are presented in the supplementary material.

We assessed four types of adaptation scenarios by using the digits datasets, namely MNIST~\cite{lecun1998gradient}, Street View House Numbers (SVHN)~\cite{netzer2011reading}, and USPS~\cite{hull1994database}. We further evaluated our method on the traffic sign datasets, Synthetic Traffic Signs (SYN SIGNS)~\cite{moiseev2013evaluation} and the German Traffic Signs Recognition Benchmark~\cite{stallkamp2011german} (GTSRB). 
In this experiment, we employed the CNN architecture used in~\cite{ganin2014unsupervised} and~\cite{bousmalis2016unsupervised}.  We added batch normalization to each layer in these models. We used Adam~\cite{kingma2014adam} to optimize our model and set the learning rate as $2.0\times10^{-4}$ in all experiments. We set the batch size to 128 in all experiments. The hyper-parameter peculiar to our method was $n$, which denotes the number of times we update the feature generator to mimic classifiers. We varied the value of $n$ from $2$ to $4$ in our experiment and observed the sensitivity to the hyper-parameter. We followed the protocol of unsupervised domain adaptation and did not use validation samples to tune hyper-parameters. The other details are provided in our supplementary material due to a limit of space.

\textbf{SVHN$\rightarrow$MNIST}

SVHN~\cite{netzer2011reading} and MNIST~\cite{lecun1998gradient} have distinct properties because SVHN datasets contain images with a colored background, multiple digits, and extremely blurred digits, meaning that the domain divergence is very large between these datasets. 

\textbf{SYN SIGNS$\rightarrow$GTSRB}
In this experiment, we evaluated the adaptation from synthesized traffic signs datasets (SYN SIGNS dataset~\cite{ganin2014unsupervised}) to real-world signs datasets (GTSRB dataset~\cite{stallkamp2011german}). These datasets contain 43 types of classes.

\textbf{MNIST$\leftrightarrow$USPS}
We also evaluate our method on MNIST and USPS datasets~\cite{lecun1998gradient} to compare our method with other methods. We followed the different protocols provided by the paper, ADDA~\cite{tzeng2017adversarial} and PixelDA~\cite{bousmalis2016unsupervised}. 

\textbf{Results}
Table \ref{tb:exp_digit} lists the accuracies for the target samples, and Fig. \ref{fig:svhn_graph} and \ref{fig:synth_graph} show the relationship between the discrepancy loss and accuracy during training. For the {\it source only} model, we used the same network architecture as used in our method. Details are provided in the supplementary material.
We extensively compared our methods with distribution matching-based methods as shown in Table \ref{tb:exp_digit}. The proposed method outperformed these methods in all settings. The performance improved as we increased the value of $n$. Although other methods such as ATDA~\cite{saito2017asymmetric} performed better than our method in some situations, the method utilized a few labeled target samples to decide hyper-parameters for each dataset. The performance of our method will improve too if we can choose the best hyper-parameters for each dataset. 
As Fig. \ref{fig:svhn_graph} and \ref{fig:synth_graph} show, as the discrepancy loss diminishes, the accuracy improves, confirming that minimizing the discrepancy for target samples can result in accurate adaptation.

We visualized learned features as shown in Fig. \ref{fig:gtsr_noad} and \ref{fig:gtsr_ad}.
Our method did not match the distributions of source and target completely as shown in Fig. \ref{fig:gtsr_ad}. However, the target samples seemed to be aligned with each class of source samples.
Although the target samples did not separate well in the non-adapted situation, they did separate clearly as do source samples in the adapted situation. 
\begin{figure*}[t!]
\centering
  \includegraphics[width=0.9\linewidth]{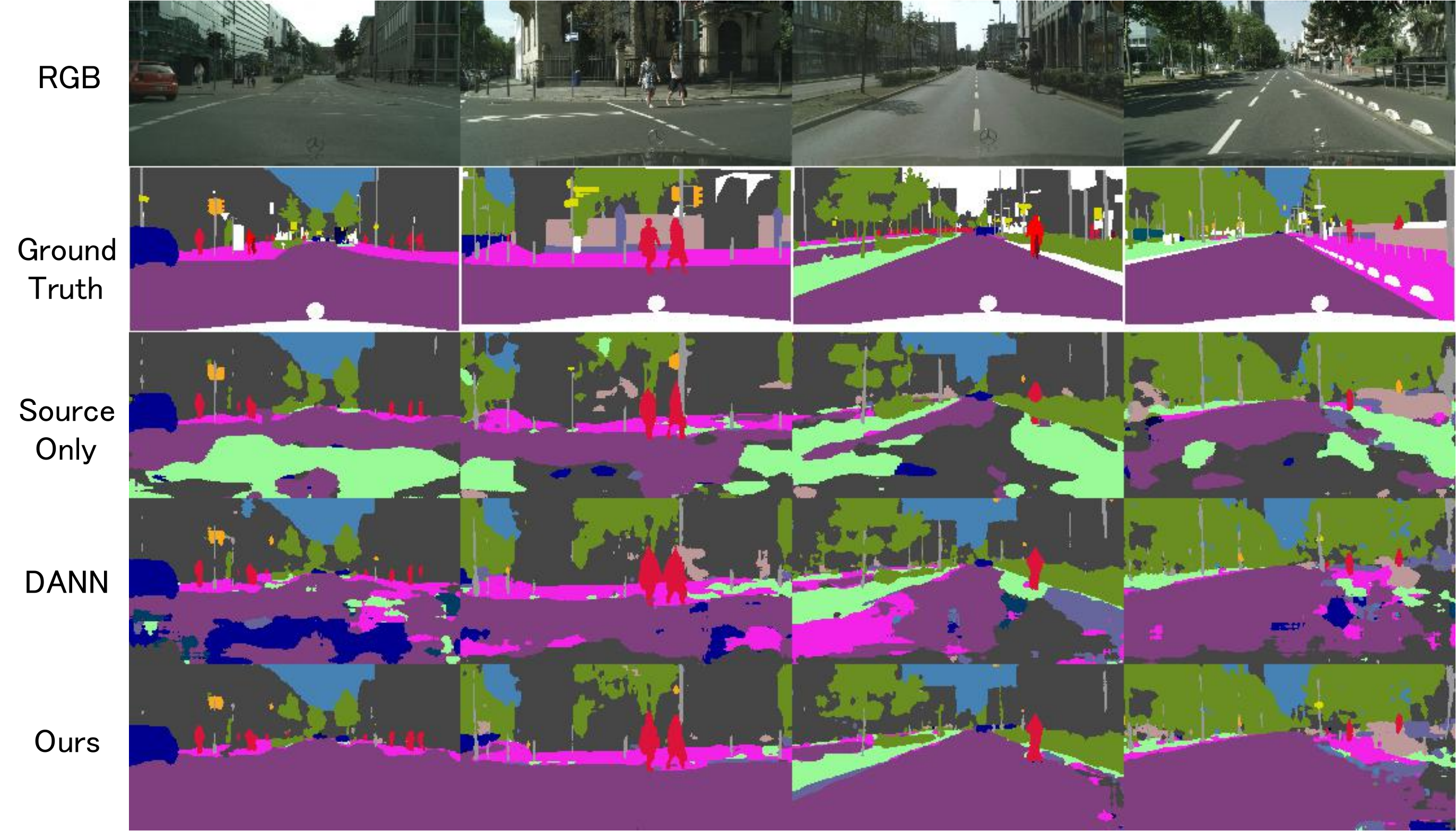}
  \caption{Qualitative results on adaptation from GTA5 to Cityscapes. DRN-105 is used to obtain these results.}
  \label{fig:vis_gta2city}
\end{figure*}

\begin{table*}[tbp]
\begin{center}
\scalebox{0.65}{
\begin{tabular}{l|l||c|ccccccccccccccccccc}
  \toprule[1.5pt]
 \scriptsize{Network}&method   & mIoU & road & sdwk & bldng & wall & fence & pole & light & sign & vgttn & trrn & sky  & person & rider & car  & truck & bus  & train & mcycl & bcycl \\
  \hline \hline

\scriptsize{VGG-16} &Source Only  & 24.9 & 25.9 & 10.9 & 50.5 & 3.3  & 12.2 & 25.4 & 28.6 & 13.0 & 78.3 & 7.3  & 63.9 & 52.1 & 7.9 & 66.3 & 5.2  & 7.8  & 0.9 & 13.7 & 0.7 \\
 & FCN Wld~\cite{hoffman2016fcns}  & 27.1 & 70.4 & 32.4 & 62.1  & 14.9 & 5.4   & 10.9 & 14.2  & 2.7  & 79.2  & 21.3 & 64.6 & 44.1   & 4.2   & 70.4 & 8.0     & 7.3  & 0.0     & 3.5   & 0.0     \\

 &  CDA (I)~\cite{Zhang_2017_ICCV}  & 23.1 & 26.4 & 10.8 & 69.7  & 10.2 & 9.4   & 20.2 & 13.6  & 14.0 & 56.9  & 2.8  & 63.8 & 31.8   & 10.6  & 60.5 & 10.9  & 3.4  & \textbf{10.9}  & 3.8   & 9.5   \\

 &Ours (k=2) & 28.0 & 87.4 & 15.4 & 75.5 & 17.4 & 9.9  & 16.2 & 11.9 & 0.6  & 80.6 & 28.1 & 60.2 & 32.5 & 0.9 & 75.4 & 13.6 & 4.8  & 0.1 & 0.7  & 0.0 \\
  
&Ours (k=3) & 27.3 & 86.0 & 10.5 & 75.1 & 20.0 & 2.9  & 19.4 & 8.4  & 0.7  & 78.4 & 19.4 & 74.8 & 23.2 & 0.3 & 74.1 & 14.3 & 10.4 & 0.2 & 0.1  & 0.0 \\
&Ours (k=4) & 28.8 & 86.4 & 8.5  & 76.1 & 18.6 & 9.7  & 14.9 & 7.8  & 0.6  & 82.8 & 32.7 & 71.4 & 25.2 & 1.1 & 76.3 & 16.1 & 17.1 & 1.4 & 0.2  & 0.0\\\hline

\scriptsize{DRN-105}&Source Only  & 22.2 & 36.4 & 14.2 & 67.4 & 16.4 & 12.0 & 20.1 & 8.7  & 0.7  & 69.8 & 13.3 & 56.9 & 37.0 & 0.4  & 53.6 & 10.6 & 3.2  & 0.2 & 0.9  & 0.0  \\
 &DANN~\cite{ganin2014unsupervised}&32.8&64.3&23.2&73.4&11.3&\textbf{18.6}&\textbf{29.0}&\textbf{31.8}&14.9&82.0&16.8&73.2&53.9&12.4&53.3&20.4&11.0&5.0&18.7&9.8\\
&Ours (k=2) & \textbf{39.7} & 90.3 & 31.0 & 78.5 & 19.7 & 17.3 & 28.6 & 30.9 & \textbf{16.1} & 83.7 & 30.0 & 69.1 & \textbf{58.5} & \textbf{19.6} & 81.5 & 23.8 & \textbf{30.0} & 5.7 & \textbf{25.7} & \textbf{14.3} \\
&Ours (k=3) & 38.9 & \textbf{90.8} & \textbf{35.6} & \textbf{80.5} & 22.9 & 15.5 & 27.5 & 24.9 & 15.1 & 84.2 & 31.8 & 77.4 & 54.6 & 17.2 & 82.0 & 21.6 & 29.0 & 1.3 & 21.8 & 5.3  \\
&Ours (k=4) & 38.1 & 89.2 & 23.2 & 80.2 & \textbf{23.6} & 18.1 & \textbf{27.7} & 25.0 & 9.3  & \textbf{84.4} & \textbf{34.6} & \textbf{79.5} & 53.2 & 16.0 & \textbf{84.1} & \textbf{26.0} & 22.5 & 5.2 & 16.7 & 4.8 \\
  \bottomrule[1.5pt]
  \end{tabular}
  }
  \end{center}
   \vspace{-3mm}
     \caption{Adaptation results on the semantic segmentation. We evaluate adaptation from GTA5 to Cityscapes dataset.}
  
      \label{tb:gta2city}
\end{table*}
\begin{table*}[t!]
\begin{center}
\scalebox{0.65}{
\begin{tabular}{l|l||c|cccccccccccccccc}
  \toprule[1.5pt]
\scriptsize{Network}&method&mIoU     & road                & sdwlk & bldng & wall & fence & pole & light & sign & vgttn & sky & prsn & ridr & car  & bus  & mcycl & bcycl      \\ \hline \hline
VGG-16 &Source Only~\cite{Zhang_2017_ICCV}    & 22.0     & 5.6  & 11.2  & 59.6 & 0.8   & 0.5  & 21.5       & 8.0 & 5.3    & 72.4  & 75.6 & 35.1 & 9.0       & 23.6    & 4.5  & 0.5 & \textbf{18.0} \\
& FCN Wld~\cite{hoffman2016fcns}   & 20.2     & 11.5 & 19.6  & 30.8 & 4.4   & 0.0  & 20.3       & 0.1 & \textbf{11.7}   & 42.3  & 68.7 & 51.2 & 3.8       & 54.0    & 3.2  & 0.2 & 0.6  \\

&CDA (I+SP)~\cite{Zhang_2017_ICCV}   & 29.0     & 65.2 & 26.1  & 74.9 & 0.1   & \textbf{0.5}  & 10.7       & 3.7 & 3.0    & 76.1  & 70.6 & 47.1 & 8.2       & 43.2    & 20.7 & 0.7 & 13.1 \\\hline
 DRN\_105 &Source Only  & 23.4     & 14.9 & 11.4  & 58.7 & 1.9   & 0.0  & 24.1       & 1.2 & 6.0    & 68.8  & 76.0 & \textbf{54.3} & 7.1       & 34.2    & 15.0 & 0.8 & 0.0  \\
&DANN~\cite{ganin2014unsupervised}&32.5&67.0&29.1&71.5&\textbf{14.3}&0.1&28.1&\textbf{12.6}&10.3	&72.7&76.7&48.3&\textbf{12.7}&62.5&11.3&2.7&0.0\\
&Ours (k=2)& 36.3 & 83.5 & 40.9 & 77.6 & 6.0 & 0.1 & 27.9 & 6.2 & 6.0 & 83.1 & \textbf{83.5} & 51.5 & 11.8 & 78.9 & 19.8 & 4.6 & 0.0   \\
&Ours (k=3)&\textbf{37.3} & 84.8 & \textbf{43.6} & \textbf{79.0} & 3.9 & 0.2 & \textbf{29.1} & 7.2 & 5.5 & \textbf{83.8} & 83.1 & 51.0 & 11.7 & 79.9 & 27.2 & \textbf{6.2} & 0.0   \\
&Ours (k=4)&37.2 & \textbf{88.1} & 43.2 & 79.1 & 2.4 & 0.1 & 27.3 & 7.4 & 4.9 & 83.4 & 81.1 & 51.3 & 10.9 & \textbf{82.1} & \textbf{29.0} & 5.7 & 0.0\\
  \bottomrule[1.5pt]
  \end{tabular}}
  \end{center}
  \vspace{-3mm}
     \caption{Adaptation results on the semantic segmentation. We evaluate adaptation from Synthia to Cityscapes dataset.}
  
      \label{tb:synthia2city}
\end{table*}
 \subsection{Experiments on VisDA Classification Dataset}
We further evaluated our method on an object classification setting. The VisDA dataset~\cite{peng2017visda} was used in this experiment, which evaluated adaptation from synthetic-object to real-object images. To date, this dataset represents the largest for cross-domain object classification, with over 280K images across 12 categories in the combined training, validation, and testing domains. The source images were generated by rendering 3D models of the same object categories as in the real data from
different angles and under different lighting conditions. It contains 152,397 synthetic images. The validation images were collected from MSCOCO~\cite{lin2014microsoft} and they amount to  55,388 in total. In our experiment, we considered the images of validation splits as the target domain and trained models in unsupervised domain adaptation settings. We evaluate the performance of ResNet101~\cite{he2016deep} model pre-trained on Imagenet~\cite{deng2009imagenet}. The final fully-connected layer was removed and all layers were updated with the same learning rate because this dataset has abundant source and target samples. We regarded the pre-trained model as a generator network and we used three-layered fully-connected networks for classification networks. The batch size was set to 32 and we used SGD with learning rate $1.0\times 10^{-3}$ to optimize the model. We report the accuracy after 10 epochs.
The training details for baseline methods are written in our supplementary material due to the limit of space.

\textbf{Results}
Our method achieved an accuracy much better than other distribution matching based methods (Table \ref{tb:visda}). In addition, our method performed better than the source only model in all classes, whereas MMD and DANN perform worse than the source only model in some classes such as car and plant. We can clearly see the clear effectiveness of our method in this regard. In this experiment, as the value of $n$ increase, the performance improved. We think that it was because of the large domain difference between synthetic objects and real images. The generator had to be updated many times to align such distributions.

\section{Experiments on Semantic Segmentation}
We further applied our method to semantic segmentation. Considering a huge annotation cost for semantic segmentation datasets, adaptation between different domains is an important problem in semantic segmentation. 

\textbf{Implementation Detail}
We used the publicly available synthetic dataset GTA5~\cite{richter2016playing} or Synthia~\cite{ros2016synthia} as the source domain dataset and real dataset Cityscapes~\cite{cordts2016cityscapes} as the target domain dataset.
Following the work~\cite{hoffman2016fcns,Zhang_2017_ICCV}, the Cityscapes validation set was used as our test set. As our training set, the Cityscapes train set was used.
During training, we randomly sampled just a single sample (setting the batch size to 1 because of the GPU memory limit) from both the images (and their labels) of the source dataset and the remaining images of the target dataset but with no labels. 

We applied our method to VGG-16~\cite{simonyan2014very} based FCN-8s~\cite{long2015fully} and DRN-D-105~\cite{Yu2017} to evaluate our method.
The details of models, including their architecture and other hyper-parameters, are described in the supplementary material.

We used Momentum SGD to optimize our model and set the momentum rate to 0.9 and the learning rate to $1.0 \times 10^{-3}$ in all experiments.
The image size was resized to $1024 \times 512$.
Here, we report the output of $F_1$ after 50,000 iterations.

\textbf{Results}
Table \ref{tb:gta2city}, Table \ref{tb:synthia2city}, and Fig. \ref{fig:vis_gta2city} show quantitative and qualitative results, respectively.
These results illustrate that even with a large domain difference between synthetic to real images, our method is capable of improving the performance. Considering the mIoU of the model trained only on source samples, we can see the clear effectiveness of our adaptation method. Also, compared to the score of DANN, our method shows clearly better performance.

   \section{Conclusion}
     \vspace{-1mm}
In this paper, we proposed a new approach for UDA, which utilizes task-specific classifiers to align distributions. We propose to utilize task-specific classifiers as discriminators that try to detect target samples that are far from the support of the source. A feature generator learns to generate target features near the support to fool the classifiers. Since the generator uses feedback from task-specific classifiers, it will avoid generating target features near class boundaries.
We extensively evaluated our method on image classification and semantic segmentation datasets. In almost all experiments, our method outperformed state-of-the-art methods. We provide the results when applying gradient reversal layer~\cite{ganin2014unsupervised} in the supplementary material, which enables to update parameters of the model in one step. 
\section{Acknowledgements}
The work was partially supported by CREST, JST, and was partially funded by the ImPACT Program of the Council for Science, Technology, and Innovation
(Cabinet Office, Government of Japan).
{\small
\bibliographystyle{ieee}
\bibliography{egbib}
}

We would like to show supplementary information for our main paper.
First, we introduce the detail of the experiments. Finally, we show some additional results of our method. 
\section*{Toy Dataset Experiment}
We show the detail of experiments on toy dataset in main paper. 
\subsection*{Detail on experimental setting}
The detail of experiment on toy dataset is shown in this section. When generating target samples, we set the rotation angle 30 in experiments of our main paper. 
We used Adam with learning rate $2.0\times10^{-4}$ to optimizer the model. The batch size was set to 200. For a feature generator, we used 3-layered fully-connected networks with 15 neurons in hidden layer, in which ReLU is used as the activation function. For classifiers, we used three-layed fully-connected networks with 15 neurons in hidden layer and 2 neurons in output layer. The decision boundary shown in the main paper is obtained when we rotate the source samples 30 degrees to generate target samples. We set $n$ to $3$ in this experiment. 

\section*{Experiment on Digit Dataset}
We report the accuracy after training 20,000 iterations except for the adaptation between MNIST and USPS. Due to the lack of training samples of the datasets, we stopped training after 200 epochs (13 iterations per one epoch) to prevent over-fitting.
We followed the protocol presented by \cite{ganin2014unsupervised} in the following three adaptation scenarios.
\textbf{SVHN$\rightarrow$MNIST}
In this adaptation scenario, we used the standard training set as training samples, and testing set as testing samples both for source and target samples.

\textbf{SYN DIGITS$\rightarrow$SVHN}
We used 479400 source samples and 73257 target samples for training, 26032 samples for testing. 

\textbf{SYN SIGNS$\rightarrow$GTSRB}
We randomly selected 31367 samples for target training and evaluated the accuracy on the rest. 

\textbf{MNIST$\leftrightarrow$USPS}
In this setting, we followed the different protocols provided by the paper, ADDA~\cite{tzeng2017adversarial} and PixelDA~\cite{bousmalis2016unsupervised}. The former protocol provides the setting where a part of training samples are utilized during training. 2,000 training samples are picked up for MNIST and 1,800 samples are used for USPS.
The latter one allows to utilize all training samples during training. We utilized the architecture used as a classification network in PixelDA~\cite{bousmalis2016unsupervised}. We added Batch Normalization layer to the architecture.

\section*{Experiment on VisDA Classification Dataset}
The detail of architecture we used and the detail of other methods are shown in this section.

\textbf{Class Balance Loss}
In addition to feature alignment loss, we used a class balance loss to improve the accuracy in this experiment. Please note that we incorporated this loss in comparable methods too. We aimed to assign the target samples to each classes equally. Without this loss, the target samples can be aligned in an unbalanced way. 
The loss is calculated as follows:
\begin{equation}
  {\mathbb{E}_{\mathbf{x_{t}}\sim X_{t}}}\sum_{k=1}^{K} \log p(y=k|{\mathbf x_{t}})\\
  \end{equation}
The constant term $\lambda=0.01$ was multiplied to the loss and add this loss in Step 2 and Step 3 of our method. This loss was also introduced in MMD and DANN too when updating parameters of the networks. 

For the fully-connected layers of classification networks, we set the number of neurons to $1000$. In order to fairly compare our method with others, we used the exact the same architecture for other methods. 

\textbf{MMD}
We calculated the maximum mean discrepancy (MMD) \cite{long2015learning}, namely the last layer of feature generator networks. We used RBF kernels to calculate the loss.
We used the the following standard deviation parameters:
\begin{equation}
  \sigma = [0.1,0.05,0.01,0.0001,0.00001]
\end{equation}
We changed the number of the kernels and their parameters, but we could not observe significant performance difference. We report the performance after 5 epochs. We could not see any improvement after the epoch.

\textbf{DANN}
To train a model~(\cite{ganin2014unsupervised}), we used two-layered domain classification networks. We set the number of neurons in the hidden layer as $100$. We also used Batch Normalization, ReLU and dropout layer. Experimentally, we did not see any improvement when the network architecture is changed. According to the original method~(\cite{ganin2014unsupervised}), learning rate is decreased every iteration. However, in our experiment, we could not see improvement, thus, we fixed learning rate $1.0\times10^{-3}$. In addition, we did not introduce gradient reversal layer for our model. We separately update discriminator and generator. We report the accuracy after 1 epoch.

\section*{Experiments on Semantic Segmentation}
We describe the details of our experiments on semantic segmentation.

\subsection*{Details}

\textbf{Datasets}
GTA~\cite{richter2016playing}, Synthia~\cite{ros2016synthia} and Cityscapes~\cite{cordts2016cityscapes} are vehicle-egocentric image datasets but GTA and Synthia are synthetic and Cityscapes is real world dataset.
GTA is collected from the open world in the realistically rendered computer game Grand Theft Auto V (GTA, or GTA5).
It contains 24,996 images, whose semantic segmentation annotations are fully compatible with the classes used in Cityscapes.
Cityscapes is collected in 50 cities in Germany and nearby countries. We only used dense pixel-level annotated dataset collected in 27 cities. It contains 2,975 training set, 500 validation set, and 1525 test set. We used training and validation set. Please note that the labels of Cityscapes are just used for evaluation and never used in training. 
Similarly, we used the training splits of Synthia dataset to train our model. 

\textbf{Training Details}
When training, we ignored the pixelwise loss that is annotated \textit{backward} (\textit{void}).
Therefore, when testing, no predicted \textit{backward} label existed.
The weight decay ratio was set to $2 \times 10^{-5}$ and we used no augmentation methods.

\textbf{Network Architecture}
We applied our method to FCN-8s based on VGG-16 network. Convolution layers in original VGG-16 networks are used as generator and fully-connected layers are used as classifiers. For DRN-D-105, we followed the implementation of \url{https://github.com/fyu/drn}.
We applied our method to dilated residual networks \cite{Yu2016,Yu2017} for base networks. We used {\it DRN-D-105} model. We used the last convolution networks as classifier networks. All of lower layers are used as a generator. 

\textbf{Evaluation Metrics}
As evaluation metrics, we use intersection-over-union (IoU) and pixel accuracy.
We use the evaluation code\footnote{https://github.com/VisionLearningGroup/taskcv-2017-public/blob/master/segmentation/eval.py} released along with VisDA challenge \cite{peng2017visda}.
It calculates the PASCAL VOC intersection-over-union, i.e., $\textrm{IoU} = \frac{\textrm{TP}}{\textrm{TP}+\textrm{FP}+\textrm{FN}}$, where TP, FP, and FN are the numbers of true positive, false positive, and false negative pixels, respectively, determined over the whole test set. For further discussing our result, we also compute pixel accuracy, $\textrm{pixelAcc.} = \frac{\Sigma_{i} n_{ii}}{\Sigma_{i} t_{i}}$, where $n_{ii}$ denotes number of pixels of class $i$ predicted to belong to class $j$ and $t_{i}$ denotes total number of pixels of class $i$ in ground truth segmentation.

\section*{Additional Results}
\subsection*{Training via Gradient Reversal Layer}
In our main paper, we provide the training procedure that consists of three training steps and the number of updating generator ($k$) is a hyper-parameter in our method. We found that introducing gradient reversal layer (GRL)~\cite{ganin2014unsupervised} enables to update our model in only one step and works well in many settings. This improvement makes training faster and deletes hyper-parameter in our method. We provide the detail of the improvement and some experimental results here. 

\textbf{Training Procedure}
We simply applied gradient reversal layer when updating classifiers and generator in an adversarial manner. The layer flips the sign of gradients when back-propagating the gradient. Therefore, update for maximizing the discrepancy via classifier and minimizing it via generator was conducted simultaneously. We publicize the code with this implementation.

\begin{figure*}[t!]
  \includegraphics[width=\linewidth]{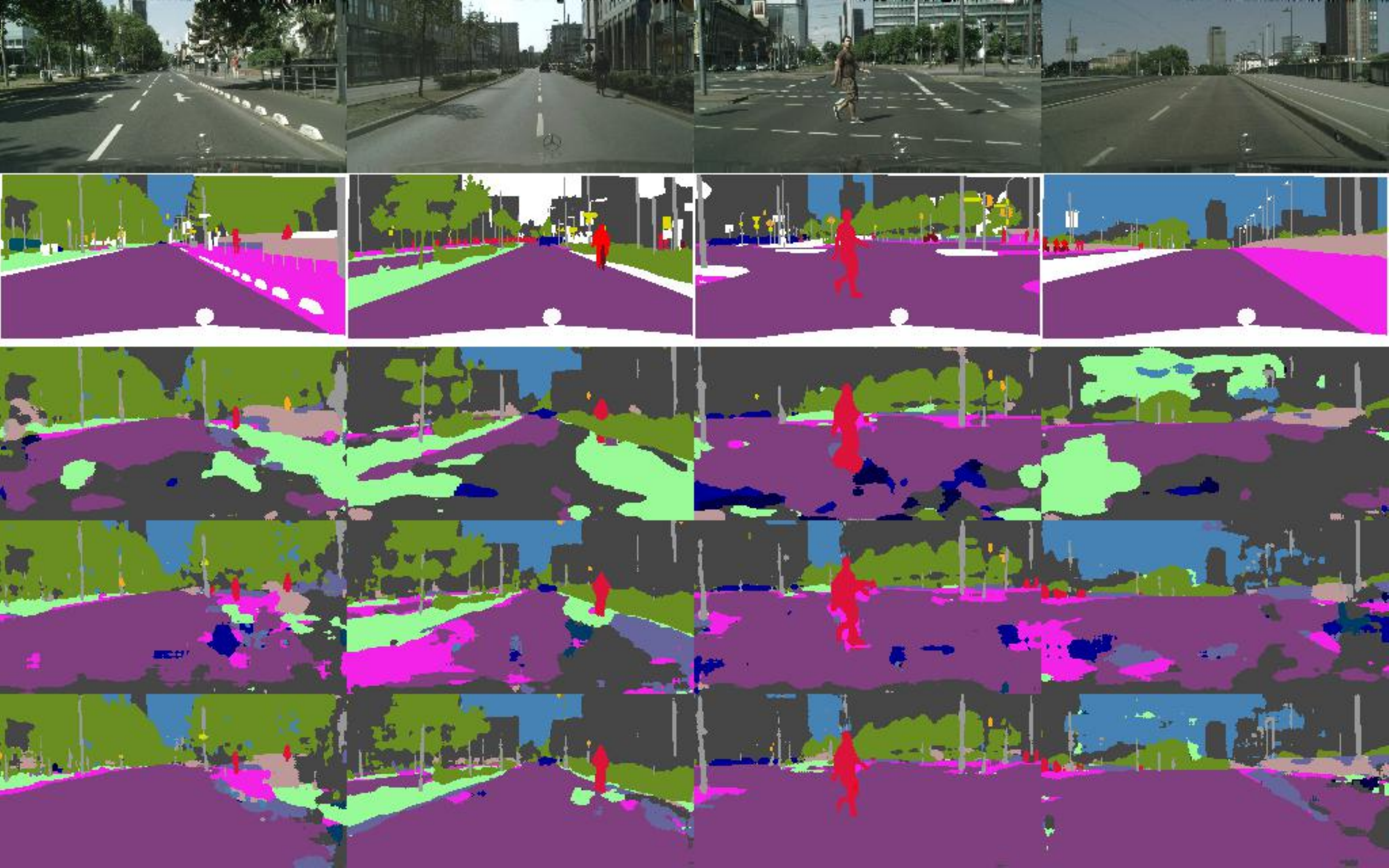}
  \caption{Qualitative results on adaptation from GTA5 to Cityscapes. From top to bottom, input, ground truth, result of source only model, DANN, and our proposed method.}
  \label{fig:vis_gta2city_sup}
\end{figure*}

\textbf{Results}
The experimental results on semantic segmentation are shown in Table \ref{tb:gta2city_sup},\ref{tb:synthia2city_sup}, and Fig. \ref{fig:vis_gta2city_sup}. Our model with GRL shows the same level of performance compared to the model trained with our proposed training procedure. 

\subsection*{Sensitivity to Hyper-Parameter}
The number of updating generator is the hyper-parameter peculiar to our method. Therefore, we show additional experimental results related to it. 
We employed the adaptation from SVHN to MNIST and conducted experiments where $n=5,6$. The accuracy was 96.0\% and 96.2\% on average. The accuracy seems to increase as we increase the value though it saturates. Training time required to obtain high accuracy can increase too. However, considering the results of GRL on semantic segmentation, the relationship between the accuracy and the number of $n$ seems to depend on which datasets to adapt.
\begin{table*}[tbp]
\begin{center}
\scalebox{0.65}{
\begin{tabular}{l|l||c|ccccccccccccccccccc}
  \toprule[1.5pt]
 \scriptsize{Network}&method   & mIoU & road & sdwk & bldng & wall & fence & pole & light & sign & vgttn & trrn & sky  & person & rider & car  & truck & bus  & train & mcycl & bcycl \\
  \hline \hline
\scriptsize{VGG-16} &Source Only  & 24.9 & 25.9 & 10.9 & 50.5 & 3.3  & 12.2 & 25.4 & 28.6 & 13.0 & 78.3 & 7.3  & 63.9 & 52.1 & 7.9 & 66.3 & 5.2  & 7.8  & 0.9 & 13.7 & 0.7 \\
 & FCN Wld~\cite{hoffman2016fcns}  & 27.1 & 70.4 & 32.4 & 62.1  & 14.9 & 5.4   & 10.9 & 14.2  & 2.7  & 79.2  & 21.3 & 64.6 & 44.1   & 4.2   & 70.4 & 8.0     & 7.3  & 0.0     & 3.5   & 0.0     \\
 &  CDA (I)~\cite{Zhang_2017_ICCV}  & 23.1 & 26.4 & 10.8 & 69.7  & 10.2 & 9.4   & 20.2 & 13.6  & 14.0 & 56.9  & 2.8  & 63.8 & 31.8   & 10.6  & 60.5 & 10.9  & 3.4  & \textbf{10.9}  & 3.8   & 9.5   \\

 &Ours (k=2) & 28.0 & 87.4 & 15.4 & 75.5 & 17.4 & 9.9  & 16.2 & 11.9 & 0.6  & 80.6 & 28.1 & 60.2 & 32.5 & 0.9 & 75.4 & 13.6 & 4.8  & 0.1 & 0.7  & 0.0 \\
  
&Ours (k=3) & 27.3 & 86.0 & 10.5 & 75.1 & 20.0 & 2.9  & 19.4 & 8.4  & 0.7  & 78.4 & 19.4 & 74.8 & 23.2 & 0.3 & 74.1 & 14.3 & 10.4 & 0.2 & 0.1  & 0.0 \\
&Ours (k=4) & 28.8 & 86.4 & 8.5  & 76.1 & 18.6 & 9.7  & 14.9 & 7.8  & 0.6  & 82.8 & 32.7 & 71.4 & 25.2 & 1.1 & 76.3 & 16.1 & 17.1 & 1.4 & 0.2  & 0.0\\\hline
&Ours (GRL)&27.3&86.2&16.1&74.4&20.7&9.5&21.5&14.8&0.1&80.4&27.8&50.3&33.9&1.2&67.6&10.8&3.0&0.2&0.9&0.0\\\hline
\scriptsize{DRN-105}&Source Only  & 22.2 & 36.4 & 14.2 & 67.4 & 16.4 & 12.0 & 20.1 & 8.7  & 0.7  & 69.8 & 13.3 & 56.9 & 37.0 & 0.4  & 53.6 & 10.6 & 3.2  & 0.2 & 0.9  & 0.0  \\
&DANN~\cite{ganin2014unsupervised}&32.8&64.3&23.2&73.4&11.3&18.6&29.0&31.8&14.9&82.0&16.8&73.2&53.9&12.4&53.3&20.4&11.0&5.0&18.7&9.8\\
&Ours (k=2) & 39.7 & 90.3 & 31.0 & 78.5 & 19.7 & 17.3 & 28.6 & 30.9 & 16.1 & 83.7 & 30.0 & 69.1 & \textbf{58.5} & \textbf{19.6} & 81.5 & 23.8 & \textbf{30.0} & 5.7 & \textbf{25.7} & \textbf{14.3} \\
&Ours (k=3) & 38.9 & \textbf{90.8} & \textbf{35.6} & \textbf{80.5} & 22.9 & 15.5 & 27.5 & 24.9 & 15.1 & 84.2 & 31.8 & 77.4 & 54.6 & 17.2 & 82.0 & 21.6 & 29.0 & 1.3 & 21.8 & 5.3  \\
&Ours (k=4) & 38.1 & 89.2 & 23.2 & 80.2 & \textbf{23.6} & 18.1 & 27.7 & 25.0 & 9.3  & \textbf{84.4} & \textbf{34.6} & \textbf{79.5} & 53.2 & 16.0 & \textbf{84.1} & \textbf{26.0} & 22.5 & 5.2 & 16.7 & 4.8 \\\hline
&Ours (GRL)&\textbf{39.9}&90.4&34.5&79.3&20.4&\textbf{20.9}&\textbf{33.1}&28.3&\textbf{18.5}&82.4&	22.6&75.5&57.6&18.6&82.7&24.1&25.6&7.6&23.9&12.3\\
 \bottomrule[1.5pt]
  \end{tabular}
  }
  \end{center}
   \vspace{-3mm}
     \caption{Adaptation results on the semantic segmentation. We evaluate adaptation from GTA5 to Cityscapes dataset.}
  
      \label{tb:gta2city_sup}
\end{table*}
\begin{table*}[t!]
\begin{center}
\scalebox{0.65}{
\begin{tabular}{l|l||c|cccccccccccccccc}
  \toprule[1.5pt]
\scriptsize{Network}&method&mIoU     & road                & sdwlk & bldng & wall & fence & pole & light & sign & vgttn & sky & prsn & ridr & car  & bus  & mcycl & bcycl      \\ \hline \hline
VGG-16 &Source Only~\cite{Zhang_2017_ICCV}    & 22.0     & 5.6  & 11.2  & 59.6 & 0.8   & 0.5  & 21.5       & 8.0 & 5.3    & 72.4  & 75.6 & 35.1 & 9.0       & 23.6    & 4.5  & 0.5 & 18.0 \\
& FCN Wld~\cite{hoffman2016fcns}   & 20.2     & 11.5 & 19.6  & 30.8 & 4.4   & 0.0  & 20.3       & 0.1 & 11.7   & 42.3  & 68.7 & 51.2 & 3.8       & 54.0    & 3.2  & 0.2 & 0.6  \\

&CDA (I+SP)~\cite{Zhang_2017_ICCV}   & 29.0     & 65.2 & 26.1  & 74.9 & 0.1   & 0.5  & 10.7       & 3.7 & 3.0    & 76.1  & 70.6 & 47.1 & 8.2       & 43.2    & 20.7 & 0.7 & 13.1 \\\hline
 DRN\_105 &Source Only  & 23.4     & 14.9 & 11.4  & 58.7 & 1.9   & 0.0  & 24.1       & 1.2 & 6.0    & 68.8  & 76.0 & 54.3 & 7.1       & 34.2    & 15.0 & 0.8 & 0.0  \\
&DANN~\cite{ganin2014unsupervised}&32.5&67.0&29.1&71.5&14.3&0.1&28.1&12.6&10.3	&72.7&76.7&48.3&12.7&62.5&11.3&2.7&0.0\\
&Ours (k=2)& 36.3 & 83.5 & 40.9 & 77.6 & 6.0 & 0.1 & 27.9 & 6.2 & 6.0 & 83.1 & 83.5 & 51.5 & 11.8 & 78.9 & 19.8 & 4.6 & 0.0   \\
&Ours (k=3)&37.3 & 84.8 & 43.6 & 79.0 & 3.9 & 0.2 & 29.1 & 7.2 & 5.5 & 83.8 & 83.1 & 51.0 & 11.7 & 79.9 & 27.2 & 6.2 & 0.0   \\
&Ours (k=4)&37.2 & 88.1 & 43.2 & 79.1 & 2.4 & 0.1 & 27.3 & 7.4 & 4.9 & 83.4 & 81.1 & 51.3 & 10.9 & 82.1 & 29.0 & 5.7 & 0.0\\\hline
&Ours (GRL)&34.8&74.7&35.5&75.9&6.2&0.1&29.0&	7.4	&6.1&82.9&83.4&47.8&9.2&71.7&	19.3&7.0&0.0\\
  \bottomrule[1.5pt]
  \end{tabular}}
  \end{center}
  \vspace{-3mm}
     \caption{Adaptation results on the semantic segmentation. We evaluate adaptation from Synthia to Cityscapes dataset.}
  
      \label{tb:synthia2city_sup}
\end{table*}
\end{document}